\DocumentMetadata{}
\documentclass[sigconf]{acmart}

\setcopyright{none}
\renewcommand\footnotetextcopyrightpermission[1]{} 
\makeatletter
\renewcommand\@formatdoi[1]{\ignorespaces}
\makeatother

\makeatletter
\def\@copyrightpermission{}
\makeatother

\renewcommand\footnotetextcopyrightpermission[1]{} 
\AtBeginDocument{%
  \providecommand\BibTeX{{%
    \normalfont B\kern-0.5em{\scshape i\kern-0.25em b}\kern-0.8em\TeX}}}

\acmPrice{15.00}
\acmISBN{978-1-4503-XXXX-X/18/06}

\usepackage{subfig}
\usepackage{graphicx}
\usepackage[utf8]{inputenc}
\usepackage{array}
\usepackage{nccmath}
\usepackage{amsmath,bm}
\usepackage{diagbox}
\usepackage{mathtools, nccmath, relsize}
\usepackage{enumitem}
\usepackage{makecell, multirow, tabularx}
\newcolumntype{Y}{>{\centering\arraybackslash}X}
\usepackage{threeparttable}

\PassOptionsToPackage{table,xcdraw}{xcolor}

\usepackage{makecell}
\usepackage{colortbl}

\usepackage{calc}
\newcolumntype{C}[1]{>{\centering\arraybackslash}p{#1}}%
\newlength{\mycolwidth}
\newlength{\mycolwidthep}
\newlength{\mycolwidthadd}
\newlength{\mytwocolwidthadd}
\newlength{\myfourcolwidth}
\newlength{\mythreecolwidth}
\newlength{\mytwocolwidth}
\usepackage{pifont}

\begin{document}

\title{CAND: Cross-Domain Ambiguity Inference for Early Detecting Nuanced Illness Deterioration}

\author{Lo Pang-Yun Ting}
\affiliation{%
  \institution{National Cheng Kung \\University}
  \city{}
  \country{}}
\email{lpyting@netdb.csie.ncku.edu.tw}

\author{Zhen Tan}
\affiliation{%
  \institution{Arizona State University}
  \city{}
  \country{}}
\email{ztan36@asu.edu}

\author{Hong-Pei Chen}
\affiliation{%
  \institution{National Cheng Kung \\University}
  \city{}
  \country{}}
\email{hpchen@netdb.csie.ncku.edu.tw}

\author{Cheng-Te Li}
\affiliation{%
  \institution{National Cheng Kung \\University}
  \city{}
  \country{}}
\email{chengte@ncku.edu.tw}

\author{Po-Lin Chen}
\affiliation{%
  \institution{National Cheng Kung University Hospital\\ Division of Infectious Diseases}
  \city{}
  \country{}}
\email{cplin@mail.ncku.edu.tw}

\author{Kun-Ta Chuang}
\affiliation{%
  \institution{National Cheng Kung University}
  \city{}
  \country{}}
\email{ktchuang@mail.ncku.edu.tw}

\author{Huan Liu}
\affiliation{%
  \institution{Arizona State University}
  \city{}
  \country{}}
\email{huanliu@asu.edu}


\begin{abstract}

Early detection of patient deterioration is essential for timely treatment, with vital signs like heart rates being key health indicators. Existing methods tend to solely analyze vital sign waveforms, ignoring transition relationships of waveforms within each vital sign and the correlation strengths among various vital signs. Such studies often overlook nuanced illness deterioration, which is the early sign of worsening health but is difficult to detect. In this paper, we introduce \emph{CAND}, a novel method that organizes the transition relationships and the correlations within and among vital signs as domain-specific and cross-domain knowledge. \emph{CAND} jointly models these knowledge in a unified representation space, considerably enhancing the early detection of nuanced illness deterioration. In addition, \emph{CAND} integrates a Bayesian inference method that utilizes augmented knowledge from domain-specific and cross-domain knowledge to address the ambiguities in correlation strengths. With this architecture, the correlation strengths can be effectively inferred to guide joint modeling and enhance representations of vital signs. This allows a more holistic and accurate interpretation of patient health. Our experiments on a real-world ICU dataset demonstrate that \emph{CAND} significantly outperforms existing methods in both effectiveness and earliness in detecting nuanced illness deterioration. Moreover, we conduct a case study for the interpretable detection process to showcase the practicality of \emph{CAND}. 
\end{abstract}

\maketitle
\pagestyle{plain}
\section{Introduction}\label{sec:intro}

\noindent\textbf{Background.} In intensive care units (ICUs), real-time monitoring and early detection of illness deterioration are crucial because undetected deterioration can delay treatments and increase mortality~\cite{Vasilevskis2009RelationshipBD}. This highlights the need for automated systems to detect patient deterioration in an early stage. Previous research~\cite{Ye2009TimeSS, Shaffer2017AnOO, romanovsky2005fever, raghavendran2007nursing} stress the role of vital signs, such as heart rate, as critical indicators of inflammation and essential for monitoring hospitalized patients~\cite{brekke2019value}. \textcolor{black}{These vital signs are simple and cost-effective to obtain from patients ~\cite{kellett2017make}, and they can be collected in real-time by non-invasive ways to reduce the risk of \textit{nosocomial infections}~\cite{sikora2020nosocomial}.} Studies~\cite{Wattmo2016MildVM, Isola2021AnalysisOG} have found that illness deterioration is a continuous process with distinct stages. Illness severity scores~\cite{strand2008severity} effectively represent the severity across various diseases. Increases in these scores represent the \textit{nuanced illness deterioration}, indicating the gradual decline of patients' health conditions. The \textit{nuanced illness deterioration} is usually not significant enough to cause alarms in monitoring systems. However, by detecting these subtle changes early, physicians are able to provide flexible medical interventions before the patient's condition becomes worse.

\begin{figure}
\graphicspath{{figs/}}
\vskip -1em
\begin{center}
\includegraphics[width=0.42\textwidth]{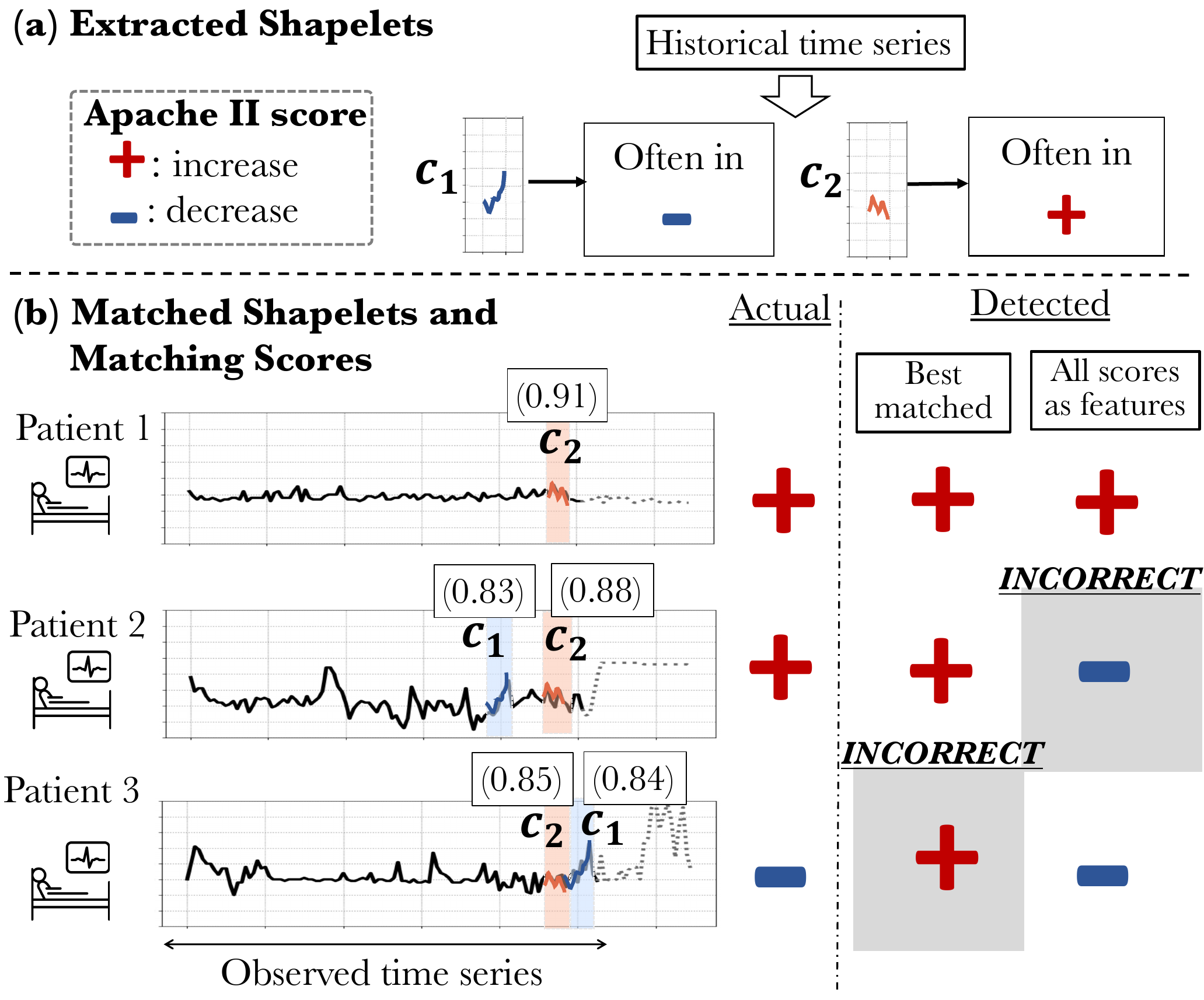}
\end{center}
\vskip -1em
\caption{The Illustration of (a) the two most critical shapelets (\textcolor{black}{$c_1$ and $c_2$}) extracted from historical heart rate time series, and (b) matched shapelets and matching scores in patients' heart rate time series. Patient 1 matches only \textcolor{black}{shapelet} $c_2$, whereas patients 2 and 3 match both shapelets with similar scores. Detections based on the best-matched shapelet or regarding all matching scores as features pose challenges in correctly detecting subtle changes in health status.}
\label{fig:patient_case}
\end{figure}

Existing methods of vital sign analysis for patient deterioration detection can fall into two categories: discovering critical information in each vital sign and understanding correlations among multiple vital signs. (\textit{\textbf{i}})~For the first category, \textcolor{black}{shapelet-based models are widely applied to analyze vital sign time series\footnote{Vital sign time series refers to sequential measurements of patients' vital signs.} to find out ``shapelets''~\cite{Ye2009TimeSS}, which represent informative waveforms or time-series subsequences. These models are able to provide interpretable results for physicians~\cite{bock2018association, Hyland2020EarlyPO, zhao2019asynchronous}}. (\textit{\textbf{ii}})~For the second category, previous works ~\cite{Fairchild2016VitalSA, Kumar2019ContinuousVS, Sullivan2021ClinicalAV} have revealed the effectiveness of analyzing multiple vital signs and their correlation in disease detection. However, these methods are generally targeted at detecting specific diseases or major deterioration events (e.g., ICU admissions and mortality, and they may fail to detect \textit{nuanced illness deterioration} in patients. This failure occurs due to two reasons: \ding{182} First, shapelet-based models, which typically rely only on shapelets' waveforms and their matching scores (similarities between shapelets and time series), may fail to capture subtle changes in health status. Our analysis (\figurename~\ref{fig:patient_case}) of real-world ICU data\footnote{The study protocol was approved by the Institutional Review Board (IRB) of NCKUH (No. B-BR-106-044 \& No. A-ER-109-027).} indicates that these models struggle to detect nuanced illness deterioration (defined as a certain increase in Apache II scores~\cite{Knaus1985apacheii}, a widely used measure of illness severity). Especially when the patient's vital sign time series highly matches multiple shapelets, which represent different illness scenarios, increasing the difficulty of accurate detection. \ding{183} Second, previous studies analyze correlations across complete time series often overlook informative shapelets and include less critical information. Additionally, patient conditions can affect vital sign responses~\cite{chester2011vital, agelink2001standardized} and further complicate the inference of correlation strengths. This may cause misinterpretations of vital sign variations and a patient's health status.

\noindent\textbf{Our Key Idea.}
To tackle these problems, we aim to effectively capture knowledge within and across vital signs measured from patients, improving early detection of nuanced illness deterioration. In \figurename~\ref{fig:patient_case}, although $c_1$ and $c_2$ match the time series from both patient 2 (deteriorating) and patient 3 (recovering), the transitions and time intervals between shapelets are different in patient 2 ($c_1 \rightarrow c_2$) and patient 3 ($c_2 \rightarrow c_1$). This illustrates the importance of analyzing \textit{transition relationships} among shapelets to identify subtle changes in a patient's health. Therefore, we structure these \textit{transition relationships} into \textit{\textbf{domain-specific knowledge}} for each type of vital sign, improving understanding of physiological changes in patients. On the other hand, we model correlations among shapelets from different vital signs to form \textit{\textbf{cross-domain knowledge}}, while each correlation has ambiguous strengths. The main task is to infer correlation strengths in cross-domain knowledge and manage their impacts on learning domain-specific knowledge. In this manner, each domain-specific knowledge can be preserved and augmented through cross-domain insights, which helps to get a comprehensive understanding of patients' vital sign information and reduce misinterpretations of health status. 

In this paper, we propose \textbf{\underline{C}}ross-Domain \textbf{\underline{A}}mbi-guity Inference for Early Detecting \textbf{\underline{N}}uanced Illness \textbf{\underline{D}}eterioration, dubbed \emph{\textbf{CAND}}, a novel approach that jointly models multiple domain-specific and cross-domain knowledge among vital signs in the representation space. Specifically, \emph{CAND} includes two key strategies: (\textit{\textbf{i}})~First, domain-specific and cross-domain knowledge are formed into different \textit{knowledge structures}. An augmentation mechanism is then applied to capture contexts influenced by correlations in these knowledge structures. (\textit{\textbf{ii}})~Second, a Bayes-based method is designed to infer the correlation strengths to quantify the strengths of connecting knowledge across vital signs. Then, the inferred strengths are used to guide the joint modeling of domain-specific and cross-domain knowledge. Hence, vital sign time series, continuously received from patients at each time interval, can be effectively represented for real-time patient status monitoring. This way, our method can enhance early detection by considering future shapelet transitions and improve reliability by modeling knowledge within and among vital signs. This helps to balance detection earliness and accuracy for nuanced illness deterioration. 

\noindent\textbf{Contributions.} Our key contributions are listed as follows:
\vspace{-0.3em}
\begin{itemize}[leftmargin=*]
    \item We develop real-time patient monitoring and introduce a representation method for early detection of nuanced illness deterioration, using low-cost vital sign data.
    \item Our \emph{CAND} effectively incorporates domain-specific and cross-domain knowledge. This incorporation improves vital sign representations and allows more accurate health interpretations.
    \item We conduct experiments on the real-world ICU dataset and show that \emph{CAND} outperforms state-of-the-art baselines. Furthermore, a case study highlights the interpretable detection process of \emph{CAND}.
\end{itemize}

\vspace{-0.35em}

\section{Related Work}

\noindent\textbf{Early Detection of Patient Deterioration.} Previous studies focus on two main categories: detecting severe events (like death, ICU admissions, or cardiopulmonary arrest) and early diagnosis of specific diseases. For severe events, most of the research utilizes Electronic Health Records (EHR). Li \textit{et al. }~\cite{Li2020DeepAlertsDL} propose a multi-task model for early detecting death and ICU admissions. El-Rashidy \textit{et al. }~\cite{ElRashidy2020IntensiveCU} design stacking ensemble classifiers for mortality prediction. Alshwaheen \textit{et al. }\cite{Alshwaheen2021ANA} design a deep genetic algorithm for similar purposes. Hartvigsen \textit{et al. }~\cite{Hartvigsen2019AdaptiveHaltingPN} propose the EARLIEST method to analyze time series data in EHR. Shamout \textit{et al. }~\cite{Shamout2020DeepIE} develop a neural network to predict deterioration from various data. Zhao \textit{et al. }~\cite{zhao2019asynchronous} and Hyland \textit{et al. }~\cite{Hyland2020EarlyPO} extract shapelets from vital signs to predict ICU admissions and circulatory failure. For specific diseases, Ghalwash \textit{et al.} extract shapelets from ECG and blood gene expression for early diagnosis of specific diseases~\cite{Ghalwash2013ExtractionOI}, and they also introduce a shapelet-based early classification method to measure temporal uncertainty~\cite{Ghalwash2014UtilizingTP}. Huang \textit{et al. }~\cite{Huang2021SnippetPN} develop a snippet policy network for early cardiovascular disease classification. However, these studies require higher costs to access data since they analyze various types of patients' data rather than just vital signs. Other methods \cite{Ghalwash2013ExtractionOI, Ghalwash2014UtilizingTP, Huang2021SnippetPN, zhao2019asynchronous} analyze only ECG signals or vital signs but only focus on detecting specific diseases or severe events. Paganelli \textit{et al. }~\cite{Paganelli2022ANS} design a self-adaptive algorithm to build an early warning system for illness severity deterioration, but this work is limited to utilizing five specific vital signs. 
Therefore, there is a lack of a general method to early detect nuanced deterioration in various patients' conditions using low-cost vital sign data.

\noindent\textbf{Knowledge Graph Alignment.} 
Fusing embeddings from different knowledge graphs (KGs) or structures through entity (node) alignments is crucial, with research emphasizing semantic matching and GNN-based models. Semantic models like MTransE~\cite{Chen2016MultilingualKG}, IPTransE~\cite{Zhu2017IterativeEA}, and BootEA~\cite{Sun2018BootstrappingEA} use TransE for relational learning, while others like JAPE~\cite{Sun2017CrossLingualEA}, AttrE~\cite{Trisedya2019EntityAB}, and MultiKE~\cite{Zhang2019MultiviewKG} enhance entity semantics with diverse methods. GNN-based models, such as GCN-Align~\cite{Wang2018CrosslingualKG} and MuGNN~\cite{Cao2019MultiChannelGN}, focus on parameter sharing and filling missing entities. AVG-GCN~\cite{ye2019vectorized}, AliNet~\cite{sun2020knowledge}, and RREA~\cite{Mao2020RelationalRE} advance embedding learning by considering entity structures and relational reflection. KE-GCN~\cite{Yu2021KnowledgeEB} and LargeGNN~\cite{Xin2022LargescaleEA} update embeddings jointly and tackle scalability. Unlike these models that treat aligned entities as identical, our approach recognizes aligned entities as distinct shapelets from different knowledge structures, emphasizing the inference of correlation strengths between them for enhanced representations. This introduces a novel method for modeling multiple knowledge structures.

\noindent\textbf{Uncertain Knowledge Graph Embedding.}
Recent interest in uncertain KGs has led to the development of methods assigning confidence scores to triplets for more precise reasoning. Existing models like UKGE~\cite{chen2019embedding}, PASSLEAF~\cite{chen2021passleaf}, BEUrRE~\cite{chen2021probabilistic}, and BGNN~\cite{liang2023knowledge}, which focus on embedding within UKGs to predict triplet confidence. These methods, however, primarily concentrate on learning a single uncertain KG. Our approach differs by jointly modeling multiple domain-specific knowledge (deterministic) and cross-domain knowledge (ambiguous), without relying on pre-defined confidence scores. We focus on inferring connectivity strengths and guiding joint learning of knowledge structures with both deterministic and ambiguous information. This offers a unique perspective in knowledge graph modeling with uncertain or ambiguous information.

\section{Preliminary}
\label{sec:prelim}

In this section, we outline the key symbols and definitions. Generally, a knowledge graph (KG) is represented as $\mathcal{G}=\{(h,r,t)|h,t\in \mathcal{E}, r\in\mathcal{R}\}$, where $\mathcal{E}$ is the entity set, $\mathcal{R}$ is the relation set, and each triplet $(h,r,t)$ indicates that the head entity $h$ is related to the tail entity $t$ through the relationship $r$. A scoring function $f(h,r,t)$ is typically defined to assess the plausibility of a triplet being a fact, based on the representation vectors of $h$, $t$, and $r$. Given its ability to represent diverse relationships with triplet structures, a KG is ideal for modeling the complex knowledge in patients' vital sign time series. This leads us to develop two specific KG-based \textit{knowledge structures} in our \emph{CAND}.

\noindent\textbf{Definition 1. (Vital Sign):} Let $\mathcal{X}$ represent a specific type of vital sign, such as heart rate. The historical dataset for vital sign $\mathcal{X}$  is denoted as $\mathcal{D}_{\mathcal{X}}=\bigl \{ T_{\mathcal{X}}^i \bigr \}_{i=1}^N$,  consisting of $N$ time series data measured from different patients. Each time series $T_{\mathcal{X}}^i \in \mathbb{R}^m$ is a sequence of real-number readings measured from a patient's 
vital sign $\mathcal{X}$ over $m$ timestamps.

\noindent\textbf{Definition 2. (Concept):} A \textit{concept} is defined as a shapelet~\cite{Ye2009TimeSS}, representing a key health indicator identified as a subsequence within a patient's vital signs time series. For vital sign $\mathcal{X}$, we employ an existing method~\cite{Grabocka2014LearningTS} for shapelet discovery to identify a set of concepts, denoted as $C_{\mathcal{X}}=\{c_1, c_2, ..., c_{|C_{\mathcal{X}}|}\}$.

\noindent\textbf{Definition 3. (Domain-Specific Knowledge Structure):} Given a dataset $\mathcal{D}_{\mathcal{X}}$, the \textit{domain-specific knowledge structure} is defined as $\mathcal{G}(\mathcal{X})=\{(c_i, \tau, c_j)|c_i, c_j \in \mathcal{C}_{\mathcal{X}}, \tau \in \mathcal{T}_{\mathcal{X}}\}$, which captures all transition relationships within each time series belonging to $\mathcal{D}_{\mathcal{X}}$. Here, $\mathcal{C}_{\mathcal{X}}$ is a set of concepts, and $\mathcal{T}_{\mathcal{X}}$ denotes a set of relations. Each relation $\tau \in \mathcal{T}_{\mathcal{X}}$ specifies a time interval range between consecutive concepts occurring in time series belonging to $\mathcal{D}_{\mathcal{X}}$. Consequently, each triplet $(c_i, \tau, c_j) \in \mathcal{G}(\mathcal{X})$ signifies that concept $c_i$ transitions to concept $c_j$ over the actual time interval $\tau$.

\begin{figure}[t]
\graphicspath{{figs/}}
\begin{center}
\includegraphics[width=0.48\textwidth]{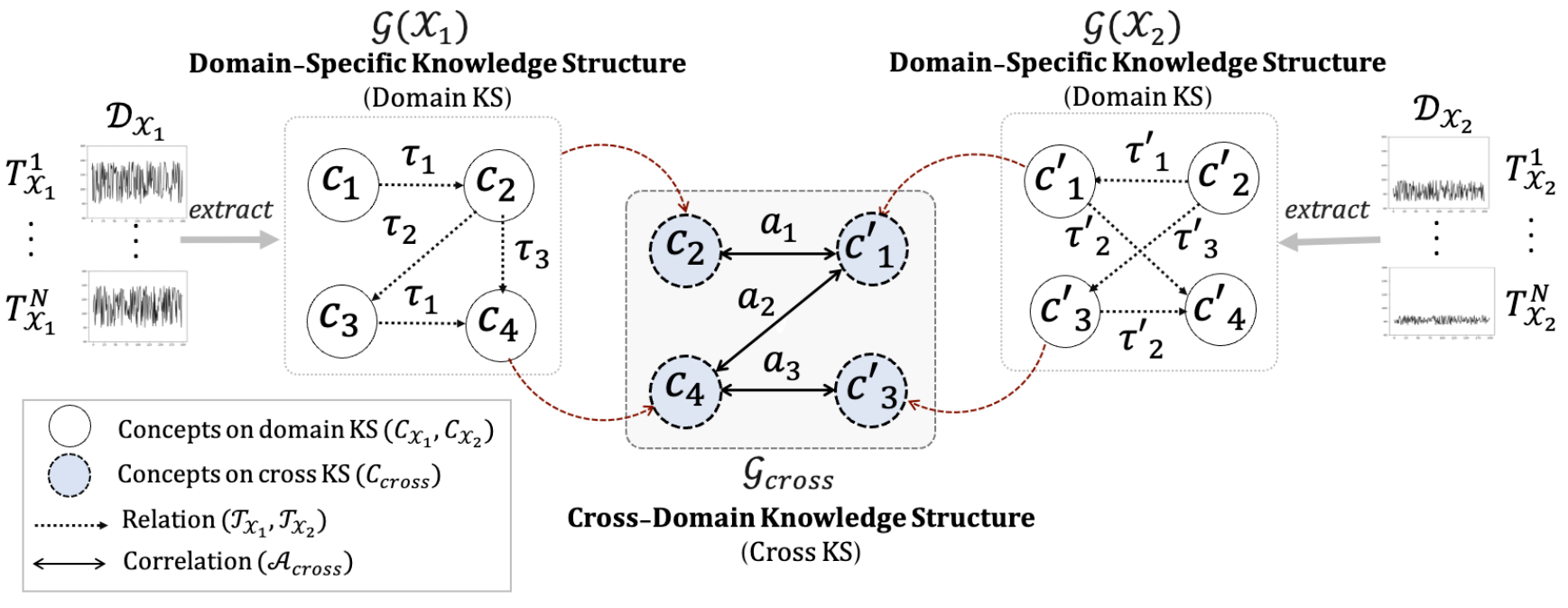}
\end{center}
\vskip -1em
\caption{The illustration of domain-specific and cross-domain knowledge structures.}
\label{fig:deisgned_KS}
\end{figure}

\noindent\textbf{Definition 4. (Cross-Domain Knowledge Structure):} Given two domain-specific knowledge structures, $\mathcal{G}(\mathcal{X}_1)$ and $\mathcal{G}(\mathcal{X}_2)$, we define a \textit{cross-domain knowledge structure} as $\mathcal{G}_{cross} = \{(c, a, c')|c, c'\in \mathcal{C}_{cross}, a \in \mathcal{A}_{cross}\}$. This structure captures correlations between concepts from vital signs $\mathcal{X}_1$ and $\mathcal{X}_2$, found in datasets $\mathcal{D}_{\mathcal{X}_1}$ and $\mathcal{D}_{\mathcal{X}_2}$, as illustrated in \figurename~\ref{fig:deisgned_KS}. The set $\mathcal{C}_{cross}$ is a subset of concepts from both domain-specific knowledge structures, denoted as $\mathcal{C}_{cross} \subset \mathcal{C}_{\mathcal{X}_1} \cup \mathcal{C}_{\mathcal{X}_2}$. The set $\mathcal{A}_{cross}$ comprises \textit{correlations} with ambiguous strengths, each correlation $a\in \mathcal{A}_{cross}$ representing a distinct likelihood range for simultaneous occurrences of $c$ and $c'$ in a patient. Each triplet $(c, a, c') \in \mathcal{G}_{cross}$ signifies a correlation $a$ between \textit{cross-domain concepts} $c$ and $c'$, with $c$ and $c'$ from $\mathcal{G}(\mathcal{X}_1)$ and $\mathcal{G}(\mathcal{X}_2)$, respectively. Note that correlations in $\mathcal{G}_{cross}$ are bidirectional, indicating that $(c, a, c')$ also implies $(c', a, c)$.

Details on constructing knowledge structures are provided in Appendix \ref{appendix:ks_construct}. Note that to distinguish between asymmetric (domain-specific) and symmetric (cross-domain) relationships in these structures, we employ the scoring function of ComplEx model ~\cite{Trouillon2016ComplexEF} in our framework.

\noindent\underline{\textbf{\textit{Early Detection for Nuanced Illness Deterioration:}}} Given two historical datasets $D_{\mathcal{X}_1}$ and $D_{\mathcal{X}_2}$, along with knowledge structures $\mathcal{G}(\mathcal{X}_1)$, $\mathcal{G}(\mathcal{X}_2)$ and $\mathcal{G}_{cross}$, we analyze currently observed vital signs $M_{1:p}=\{{T'}_{\mathcal{X}_1}, {T'}_{\mathcal{X}_2}\}$ at each testing interval $p$. Here, ${T'}_{\mathcal{X}_1}$ and ${T'}_{\mathcal{X}_2}$ represent time series data for vital signs $X_1$ and $X_2$, observed from intervals $1$ to $p$ from a patient. We aim to determine if this patient is deteriorating based on $M_{1:p}$ before certain changes in the patient's illness severity score.

Note that the graph-based architecture of \emph{CAND} facilitates its flexible extension to analyze additional types of vital signs. For simplicity, we will use the terms \textit{Domain KS} and \textit{Cross KS} to refer to \textit{Domain-Specific Knowledge Structure} and \textit{Cross-Domain Knowledge Structure}, respectively.

\begin{figure*}[t]
\graphicspath{{figs/}}
\begin{center}
\includegraphics[width=0.9\textwidth]{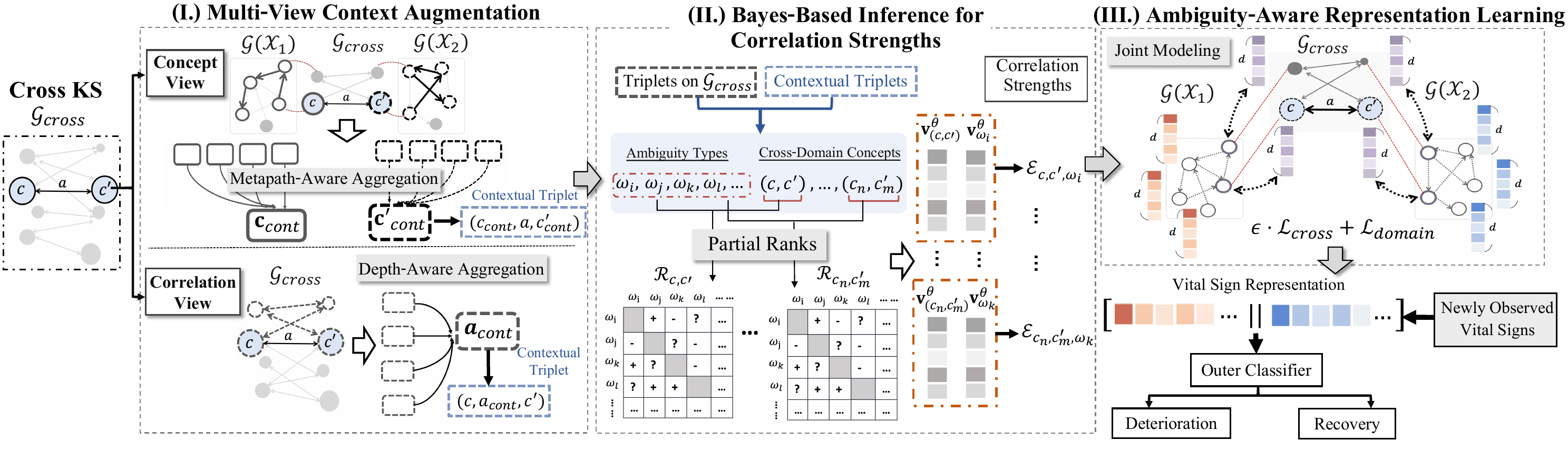}
\end{center}
\caption{The overview of the \emph{CAND} framework. (I.) A multi-view context augmentation is designed to explore specific contexts from different views. (II.) Then, a Bayes-based method is employed to infer the correlation strengths. (III.) The inferred correlation strengths then guide the joint modeling of domain KSs and the cross KS. Finally, newly observed vital signs are represented by concatenating segment vectors for outer classifiers to detect nuanced illness deterioration.}
\label{fig:framework}
\end{figure*}

\section{The \emph{CAND} Framework}

The architecture of \emph{CAND} is illustrated in \figurename~\ref{fig:framework}. \emph{CAND} comprises three main components to augment knowledge based on current knowledge structures (Sec.\ref{subsec:multi_view}) and infer strengths of correlations (Sec.\ref{subsec:bayesian}), which are then leveraged to guide joint modeling of domain KSs and the cross KS (Sec.~\ref{subsec:joint_model}).

\subsection{Multi-View Context Augmentation}
\label{subsec:multi_view}

In graph-based representation models, the contexts (or neighborhoods) of nodes are often crucial sources of information~\cite{hamilton2017inductive, Wang2020RelationalMP, sun2020knowledge}. In our cross KS, stronger correlations between cross-domain concepts indicate closely related contexts with significant mutual influence. To identify key contexts, we employ a multi-view context augmentation inspired by ~\cite{hao2019universal, wei-etal-2020-uncertainty}. The designed augmentation captures specific contexts of the cross KS from different views, serving as indicators of strengths for correlations.

For each triplet $(c,a,c')$ in cross KS $\mathcal{G}_{cross}$, we explore contexts of cross-domain concepts $(c,c')$  (Sec.~\ref{subsubsec:domain_view}) and contexts of correlation $a$ (Sec.~\ref{subsubsec:concept_view}). These contexts are formulated into \textbf{contextual triplets} to improve the modeling of the cross KS.

\vspace{-0.3em}
\subsubsection{\textit{Concept-View Contextual Triplet}}
\label{subsubsec:domain_view} 
From the concept view, we aim to explore the contexts of cross-domain concepts $(c,c')$. Instead of simply considering $n$-hop neighbors, we employ a \textbf{guided exploration}, inspired by node2vec ~\cite{grover2016node2vec}, to focus on contexts \textit{most likely to be influenced by} $c$ and $c'$  \textit{within their respective domain KSs}. Assuming $c$ belongs to $\mathcal{G}_{cross}$ and domain KS $\mathcal{G}(\mathcal{X}_1)$, we demonstrate guided exploration for $c$ in $\mathcal{G}(\mathcal{X}_1)$. Considering an exploration starting at $c$, proceeding through $w$ to $x$, the unnormalized probability of moving to the next concept $y$ is defined as follows:

\vspace{-0.2em}
\begin{equation}
\label{eq:traversal_prob}
\gamma^{x\rightarrow y} = \begin{cases}
\frac{1}{\varphi}  \cdot \mathcal{W}_{x,y}&\!\!\!, \text{ if } d^{w,y}\leq 1 \text{ or } y\in \mathcal{G}_{cross} \\
 \varphi \cdot \mathcal{W}_{x,y}&\!\!\!, \parbox[t]{5.5cm}{\text{ if }  $d^{w,y}= 2$ \text{ and} $x\in \mathcal{G}_{cross} ,\: y\notin \mathcal{G}_{cross}$ }\\
\mathcal{W}_{x,y}&\!\!\!, \text{ otherwise } 
\end{cases},
\vspace{-0.3em}
\end{equation}
where $\mathcal{W}_{x,y}$ is the probability of concept $y$ occurred after $x$ in the historical dataset $\mathcal{D}_{\mathcal{X}_1}$. $d^{w,y}$ denotes the shortest distance between $w$ and $y$ on $\mathcal{G}(\mathcal{X}_1)$. Here, $\varphi>1$ biases the exploration towards a depth-first search, prioritizing concepts that are away from $w$ and do not belong to $\mathcal{G}_{cross}$ while avoiding revisits. \textcolor{black}{The supplementary descriptions of the design of Eq.~\ref{eq:traversal_prob} are in Appendix~\ref{appendix:detail_guided}.}

Formally, the guided exploration distribution for $c$ is defined as $\mathcal{D}(y|x, c)=\gamma^{x\rightarrow y}/Z$ with $Z$ as a normalizing constant as in ~\cite{grover2016node2vec}. The guided exploration distribution for $c'$ is defined in a similar manner. Let $\psi^i_c$ and $\psi^i_{c'}$ denote the sequences of explored concepts during the $i$-th exploration of $c$ and $c'$, respectively. Note that $|\psi^i_{c}|=|\psi^i_{c'}|= L$, where $L$ is a fixed exploration length. After $N$ explorations, the \textit{concept-view} contextual triplet is formulated as follows:

\vspace{-0.5em}
\begin{equation}
\label{eq:domain_triplet_formulation}
(c_{cont}, a, c'_{cont}) = \biggl( \bigl\{\psi^i_{c} \bigr\}_{i=1}^{N}  \:,\: a,\:  \bigl\{\psi^i_{c'} \bigr\}_{i=1}^{N}  \biggr).
\end{equation}
\vspace{-0.5em}

Subsequently, a \textbf{metapath-aware aggregation} is employed to construct the representation vectors of $c_{cont}$ and $c'_{cont}$, respectively. Here, a \textit{metapath} refers to a sequence of relation types within a knowledge structure, defining a composite relationship among the relations involved in explorations. For $c_{cont}$, we categorize all exploration sequences $\bigl\{\psi^i_{c} \bigr\}_{i=1}^{N}$ of $c$ into different metapaths according to the types and order of relations encountered in $\mathcal{G}(\mathcal{X}_1)$. Taking $\phi$ as a metapath, we aggregate the representation vectors of concepts within each exploration $\psi^i_{c}$  belonging to $\phi$ to develop the overall representation of $\phi$, which is formulated as follows:

\vspace{-0.5em}
\begin{equation}
\label{eq:metapath-imp}
\bold{h}^\phi  = \frac{1}{|\phi|}  \sum_{\psi_c^i \in \phi }^{}\sum_{e_j\in \psi_c^i }^{}\alpha _{j} \bold{e}_{j}^{t};\:\: \alpha _{j} = \frac{\exp(p^{c\rightarrow e_j})}{\sum_{e_k\in \psi_c^i }^{}\exp(p^{c\rightarrow e_k})},
\end{equation}
where $e_j$ represents a concept within the exploration $\psi_c^i$, with its initial representation vector $\bold{e}^t_j\in \mathbb{R}^d$ at current epoch $t$, where $d$ is the dimensionality. The weight $\alpha_j$ is determined by the  $p^{c\rightarrow e_j}$, which represents the probability of exploring from $c$ to $e_j$ via the guided exploration distribution $\mathcal{D}(y|x, c)$. 

Let $S_{meta}$ be the set of all metapaths, the final representation vector for $c_{cont}$ is then obtained as follows:

\vspace{-0.1em}
\begin{equation}
\label{eq:domain_embed}
\bold{c} _{cont} = \sum_{\phi \in S_{meta}}^{} \frac{\bold{h}^\phi}{|S_{meta}|} .
\vspace{-0.1em}
\end{equation}

Similarly, the representation vector for $c'_{cont}$ is derived in the same manner. Following this, we can acquire the representation vectors for the \textit{concept-view} contextual triplet $(c_{cont}, a, c'_{cont})$. The representation vector of $a$ is derived from its current vector.

\subsubsection{\textit{Correlation-View Contextual Triplet}}
\label{subsubsec:concept_view}

From the correlation view, we focus on exploring the contexts of correlation $a$ \textit{within cross KS} $\mathcal{G}_{cross}$, specifically for triplet $(c, a, c')$. These contexts, embodying the \textit{path information} between $(c, c')$, are crucial for deepening our understanding of correlation $a$, as they bridge the same pair of cross-domain concepts $(c, c')$. Assuming a raw path from $c$ to $c'$ in cross KG $\mathcal{G}_{cross}$ is a sequence of concepts and correlations: $c(e_0)\overset{a_1}{\rightarrow}e_1\overset{a_2}{\rightarrow}e_2 \:...\: e_{n-1}\overset{a_l}{\rightarrow}c'(e_n)$. The corresponding \textit{correlation path} $\pi=\{a_1, a_2, ..., a_n\}$ consists of all correlations in the raw path. Denote $\Pi_{\leq L'}$ as the set of all correlation paths from $c$ to $c'$ in $\mathcal{G}_{cross}$, with each correlation path $\pi \in \Pi_{\leq L'}$ no longer than $L'$. The \textit{correlation-view} contextual triplet is then formulated as follows:
\vspace{-0.1em}
\begin{equation}
\label{eq:conept_triplet_formulation}
(c, a_{cont}, c') = \Bigl( c,\: \Pi_{\leq L'}\:,\:  c' \Bigr).
\vspace{-0.2em}
\end{equation}

Subsequently, we employ a \textbf{depth-aware aggregation} to construct the representation vector of $a_{cont}$. Let $\Pi_{=l}$ represent the subset of correlation paths from $c$ to $c'$, which comprises correlation paths with identical length $l$ ($\Pi_{=l} \subset \Pi_{\leq L'}$). We first aggregate the representation vector of each correlation that presents in $\Pi_{=l}$ to formulate the representation of subset $\Pi_{=l}$ as follows:

\vspace{-0.2em}
\begin{equation}
\label{eq:concept_path_emb}
\bold{h}^{l}= \frac{1}{|\Pi_{=l}|} \sum_{\pi\in \Pi_{=l  }}^{}\sum_{a_j\in \pi}^{}\beta _{j} \boldsymbol{a}^t_j; \:\: \beta _{j}=\frac{\exp\bigl(f(e_{j-1}, a_{j}, e_j) \bigr)}{\sum_{a_k\in \pi }^{}\exp\bigl(f(e_{k-1}, a_{k}, e_k) \bigr)}
\end{equation}
where $\boldsymbol{a}^t_j\in \mathbb{R}^d$ denotes the representation vector of correlation $a_j$ at current epoch $t$. The term $\beta_j$ reflects the weight of $a_j$. Here, $(e_{j-1}, a_j, e_j)$ represents a triplet presents in a raw path corresponding to correlation path $\pi$, and $f(\cdot)$ is a scoring function as defined in Sec.~\ref{sec:prelim}. The final representation vector for $a_{cont}$ is obtained by aggregating representations of correlation path subsets. Define $S_{path}$ as the set of correlation path subsets, where each subset $\Pi_{=l}$ with $l\leq L'$. The representation of $a_{cont}$ is formulated as follows:

\vspace{-0.3em}
\begin{equation}
\label{eq:concept_embed}
\boldsymbol{a}_{cont} = \sum_{\Pi_{=l} \in S_{path} }^{} \frac{\bold{h}^l}{|S_{path}|}.
\vspace{-0.2em}
\end{equation}

Finally, besides the intrinsic knowledge of each triplet $(c, a, c')\in \mathcal{G}_{cross}$ (origin view), we can obtain contextual triplets $(c_{cont}, a, c'_{cont})$ and $(c, a_{cont}, c')$, along with their representations vectors from concept view and correlation view, respectively.

\vspace{-0.5em}
\subsection{Bayes-Based Inference for Correlation Strengths}
\label{subsec:bayesian}

In typical knowledge graphs (KGs), triplet \textit{plausibility} indicates factual likelihood, which can be regarded as an indicator of correlation strengths. Drawing inspiration from previous works ~\cite{riedel2013relation, chen2015matrix, obamuyide2017contextual} that incorporate the  Bayesian Personalized Ranking (BPR) process ~\cite{Rendle2009BPRBP} into knowledge graphs (or knowledge bases), we utilize BPR to merge different plausibilities as \textit{implicit feedback} to infer the strengths of correlations. We define the set of all cross-domain concepts in $\mathcal{G}_{cross}$ as $\mathcal{U} \subset \mathcal{C}_{cross} \times \mathcal{C}_{cross}$, and then formulate the BPR input based on the following definitions.

\noindent\textbf{Definition 5. (Ambiguity Type):} We consider each triplet $(c, a, c') \in \mathcal{G}_{cross}$, along with its contextual triplets, as embodying different ambiguity types: the origin view, the concept view, and the correlation view with correlation $a$ in $(c, c')$. Therefore, the set of ambiguity types is defined as $ \Omega=\bigcup_{a\in \mathcal{A}_{cross}}^{}\{a_{\text{origin}}, a_{\text{concept}}, a_{\text{correlation}}\}$, where each ambiguity type is denoted as $\omega \in \Omega$.

\noindent\textbf{Definition 6. (Partial Rank):} For each pair of cross-domain concepts $(c,c')\in \mathcal{U}$, if its plausibility with ambiguity type $\omega_i\in \Omega$ is higher than with $\omega_j\in \Omega$ surpasses a predefined threshold, we denote a partial rank as $\omega_i >_{(c,c')} \omega_j$. The set with all partial ranks for $(c,c')$ is depicted as $\mathcal{R}_{(c,c')}=\{(\omega_i, \omega_j)| \omega_i >_{(c,c')} \omega_j\}$.

Our objective is to rank all ambiguity types for each cross-domain concept pair, learning latent factors to estimate correlation strengths. To achieve this, we aim to maximize the posterior probability $\mathcal{P}(\theta|>_{c,c'})\propto \mathcal{P}(>_{c,c'}|\theta)\mathcal{P}(\theta)$, where $>_{c,c'}$ is total ranking for $(c,c')$, and $\theta$ denotes the parameter vectors of matrix factorization (MF). We use BPR to minimize the negative log-likelihood, derived as follows:

\begin{equation}
\label{eq:loss_prior}
\begin{aligned}
&\hspace*{-0.25cm}
\mathcal{L}_{\text{infer}} = -\ln \mathcal{P}(>_{c,c'}|\theta)\mathcal{P}(\theta)
\\
&\hspace*{0.55cm}
=\hspace{-6pt}\sum_{(c,c') \in \mathcal{U}}^{}\!\sum_{(\omega_i, \omega_j) \in \mathcal{R}_{(c,c')}}^{} \hspace{-17pt} - 
\ln \Bigl(\sigma (v_{c,c',\omega_i} - v_{c,c',\omega_j}) \Bigr) + \lambda _{\theta}\left \|  \theta\right \|^2,
\end{aligned}
\vspace{-0.1em}
\end{equation}
where $\sigma(\cdot)$ is the sigmoid function. $v_{c,c',\omega_i}$ is expressed as $v_{c,c',\omega_i}:=\mathbf{v}^{\theta}_{(c,c')} \cdot \mathbf{v}^{\theta}_{\omega_i}$ based on MF, where $\mathbf{v}^{\theta}_{(c,c')}\in \mathbb{R}^d$ and $\mathbf{v}^{\theta}_{\omega_i}\in \mathbb{R}^d$ are latent factors of cross-domain concepts$(c,c')$ and ambiguity type $\omega_i$, respectively. $\lambda _{\theta}$ are regularization parameters.

Finally, we derive latent factors $\mathbf{v}^{\theta}_{(c,c')}$ and $\mathbf{v}^{\theta}_{\omega}$ for each $(c,c')\in \mathcal{U}$ and $\omega \in \Omega$. The correlation strength of ambiguity type $\omega$ with $(c,c')$ is formulated as follows:

\vspace{-0.3em}
\begin{equation}
\label{eq:implicit_influence}
\begin{aligned}
\mathcal{E}_{c,c',\omega}=\frac{\exp(\mathbf{v}^{\theta}_{(c,c')}\cdot \mathbf{v}^{\theta}_{\omega})}{\sum_{\omega_j\in \Omega}^{} \exp(\mathbf{v}^{\theta}_{(c,c')}\cdot \mathbf{v}^{\theta}_{\omega_{j}})},
\end{aligned}
\vspace{-0.2em}
\end{equation}

\subsection{Ambiguity-Aware Representation Learning}
\label{subsec:joint_model}

In this subsection, we model the cross KS with inferred correlation strengths (Eq.~\ref{eq:implicit_influence}), which then refining representations and influencing domain KS training.

\noindent\textbf{\textit{Cross-Domain Modeling.}}
Typically, knowledge graph embedding (KGE) aims to score positive triplets higher than negative ones. Let $\nu$ be a triplet on cross KS $\mathcal{G}_{\text{cross}}$ or a contextual triplet, we adopt the sigmoid loss as suggested by ~\cite{Sun2018RotatEKG}, commonly used in recent high-performing KGE models. The loss for $\nu$ is defined as follows:

\vspace{-0.3em}
\begin{equation}
\label{eq:context_triplet_loss}
\begin{aligned}
\mathcal{L}(\nu)=-\log\sigma\biggl(\gamma_{\nu}-d(\nu)\biggr) 
- \sum_{i=1}^{n}\biggl( \frac{1}{n}\cdot \log\sigma \Bigl(d(\bar{\nu}_i ) - \gamma_{\nu}\Bigr) \biggr),
\end{aligned}
\vspace{-0.5em}
\end{equation}
where $\sigma(\cdot)$ is the sigmoid function, and $\bar{\nu}_i$ represents the $i^{th}$ negative triplet. The function $d(\cdot)=-f(\cdot)$, a negation of the scoring function $f(\cdot)$, serves as the distance function for a triplet. Each negative sample with equal importance $1/n$.

To reveal the impact of correlation strengths in the triplet loss, we introduce $\gamma_\nu$ as a \textit{dynamic margin} for each triplet $\nu$. This margin adjusts the minimum distance between negative and positive triplets. Assuming triplet $\nu$ corresponds to ambiguity type $\omega\in \Omega$ (Def. 6) embodied in cross-domain concepts $(c,c')\in \mathcal{U}$, its correlation strength $\mathcal{E}_{c,c',\omega}$ (Eq.~\ref{eq:implicit_influence}) influences the dynamic margin $\gamma_\nu$. A lower $\mathcal{E}_{c,c',\omega}$ diminishes $\gamma_\nu$, reducing the impact of the correlation on the representation vectors of $c$ and $c'$ from their domain KSs. The formulation of $\gamma_\nu$ is as follows:

\vspace{-0.3em}
\begin{equation}
\label{eq:margin_adjust}
\begin{aligned}
\gamma_{\nu}=\gamma\cdot \exp\Bigl((\mathcal{E}_{c,c',\omega}-1)\cdot \xi\Bigr),
\end{aligned}
\vspace{-0.2em}
\end{equation}
where $\gamma$ is a fixed default margin, and $\xi$ is a scale factor.

Finally, the objective function for modeling the cross KS is derived by aggregating losses across all triplets in $\mathcal{G}_{cross}$ and their contextual triplets.  Defining $\mathcal{V}$ as the set of triplets in $\mathcal{G}_{cross}$ and $\mathcal{V}_{cont}(\nu)$ as the contextual triplet set for each $\nu \in \mathcal{V}$, we formulate the cross-domain loss as follows:

\begin{equation}
\label{eq:loss_cross}
\begin{aligned}
&\hspace*{-0.2cm}
\mathcal{L}_{\text{cross}}=\frac{1}{|\mathcal{V}|}\sum_{\nu\in \mathcal{V}}^{} \biggl\{ \underbrace{\mathcal{L}(\nu)}_{\text{ origin loss}} + 
\\
&\hspace*{1.0cm}
\lambda_{c} \cdot \underbrace{\frac{1}{|\mathcal{V}_{cont}(\nu)|} \cdot \Bigl[ \sum_{\nu'\in \mathcal{V}_{cont}(\nu)}^{} \mathcal{L}(\nu')\Bigr]}_{\text{contextual loss}} \biggr\} + \underbrace{\mathcal{L}_{\text{infer}}}_{\text{inference loss}},
\end{aligned}
\vspace{-0.5em}
\end{equation}
where parameter $\lambda_{c}\in [0,1]$ controls the contribution of modeling the contextual triplet.

\noindent\textbf{\textit{Domain Modeling.}}
Using Eq. \ref{eq:loss_cross}, we obtain the representation vector of each concept $c\in \mathcal{G}_{cross}$ at each epoch, which then serves as the default representation for training domain KSs. For modeling a domain KS like $\mathcal{G}(\mathcal{X}_1)$, we employ an objective function similar to Eq. \ref{eq:context_triplet_loss}. With $\mathcal{S}$ as the triplet set in $\mathcal{G}(\mathcal{X}_1)$, the objective function for modeling $\mathcal{G}(\mathcal{X}_1)$ is derived as follows:

\vspace{-0.5em}
\begin{equation}
\label{eq:loss_spec}
\begin{aligned}
&\hspace*{-0.1cm}
\mathcal{L}_{\mathcal{G}(\mathcal{X}_1)}\!=\!\frac{1}{|\mathcal{S}|}\sum_{s\in \mathcal{S}}^{}\biggl(-\text{log}\sigma \Bigl( \gamma -d(s)\Bigr)\! - \!\!\sum_{k=1}^{n} p(\bar{s}_k) \cdot \text{log} \sigma \Bigl( d(\bar{s}_k) - \gamma\Bigr) \biggr),
\\
&\hspace*{-0.1cm}
p(\bar{s}_k)=\frac{\exp \; \alpha \cdot\bigl( f(\bar{s}_k) \bigr)}{\sum_{l=1}^{n}\exp \; \alpha \cdot\bigl(  f(\bar{s}_l) \bigr)},
\vspace{-0.7em}
\end{aligned}
\end{equation}
where $d(\cdot)$ is a triplet's distance function as in Eq. \ref{eq:context_triplet_loss}. For a domain KS without ambiguities, we fix the margin value $\gamma$ and define $p(\bar{s}_k)$ as the weight for the $k^{th}$ negative sample, emphasizing the most informative ones ~\cite{Sun2018RotatEKG}. $f(\cdot)$ is the scoring function. The hyperparameter $\alpha$ is the temperature of self-adversarial weight ~\cite{Sun2018RotatEKG}.

The objective function for domain KS $\mathcal{G}(\mathcal{X}_2)$ is formulated similarly to Eq. \ref{eq:loss_spec}. Therefore, the domain loss is derived as

\vspace{-0.1em}
\begin{equation}
\label{eq:loss_domain}
\mathcal{L}_{\text{domain}}=\mathcal{L}_{\mathcal{G}(\mathcal{X}_1)} + \mathcal{L}_{\mathcal{G}(\mathcal{X}_2)}.
\end{equation}

\noindent\textbf{\textit{Joint Loss.}}
Finally, combining the cross-domain loss (Eq. \ref{eq:loss_cross}) and domain loss (Eq. \ref{eq:loss_domain}), the final objective function of the \emph{CAND} framework is formulated as follows:

\vspace{-0.1em}
\begin{equation}
\label{eq:lost_mila}
\mathcal{L}_{\emph{CAND}} = \epsilon  \cdot \mathcal{L}_{\text{cross}} + \mathcal{L}_{\text{domain}},
\end{equation}
where $\epsilon  > 0$ is a hyperparameter balancing $\mathcal{L}_{\text{cross}}$ and $\mathcal{L}_{\text{domain}}$. We alternately optimize these losses: $\theta^{\text{new}}\longleftarrow \theta^{\text{old}}-(\epsilon \eta)\bigtriangledown \mathcal{L}_{\text{cross}}$ and $\theta^{\text{new}}\longleftarrow \theta^{\text{old}}-\eta\bigtriangledown \mathcal{L}_{\text{domain}}$ in successive steps each epoch. The learning rate $\eta$ is adjusted by $\epsilon$ for cross-domain and domain losses, with the Adam optimizer ~\cite{Kingma2014AdamAM} optimizing the joint loss.

Finally, we acquire representation vectors for concepts and relations. These vectors are then utilized to represent vital sign time series by concatenating the vectors of observed concepts and relations, forming input features for classifier training. In the testing phase, newly observed time series are similarly processed, allowing real-time detection for nuanced illness deterioration. Details of the formulation of vital sign representations are in Appendix \ref{appendix:represent_timeseries}.

\section{Experiments}

\begin{table*}[!ht]
\small
\centering
\caption{\textcolor{black}{Performance comparison (in \%) of early deterioration detection. The best and second-best results are in \textbf{bold} and \underline{underlined}, respectively. Cells in \textcolor{gray}{gray} indicate earliness scores (Ear.) reach or exceed the upper quartile among methods.}}
\vspace{-1.3em}
\renewcommand{\arraystretch}{0.45}
\setlength{\tabcolsep}{0.6pt}
\begin{tabular}{lC{\mycolwidth}C{\mycolwidth}C{\mycolwidth}C{\mycolwidth}C{\mycolwidth}C{\mycolwidth}C{\mycolwidth}C{\mycolwidth}C{\mycolwidth}C{\mycolwidth}C{\mycolwidth}C{\mycolwidth}C{\mycolwidth}C{\mycolwidth}C{\mycolwidth}C{\mycolwidthadd}}
\toprule[1.3pt]
\multirow{2}{*}{Methods} & \multicolumn{5}{|C{\myfourcolwidth}}{$\quad\quad\:$ Pruning = 0\%  }  & \multicolumn{5}{|C{\myfourcolwidth}}{$\quad\quad\:$ Pruning = 30\%  }  & \multicolumn{5}{|C{\myfourcolwidth}|}{$\quad\quad\:$ Pruning = 50\% } & Overall \\ \cmidrule(rl){2-6} \cmidrule(rl){7-11} \cmidrule(rl){12-16} \cmidrule(rl){17-17} 
 & \multicolumn{1}{|C{\mycolwidth}}{Acc.   ($\uparrow$)}  &  \multicolumn{1}{C{\mycolwidth}}{Rec.   ($\uparrow$)}  & \multicolumn{1}{C{\mycolwidth}}{F1.   ($\uparrow$)} & \multicolumn{1}{C{\mycolwidthep}}{AUC ($\uparrow$)} & \multicolumn{1}{C{\mycolwidth}}{Ear.   ($\uparrow$)} & \multicolumn{1}{|C{\mycolwidth}}{Acc. ($\uparrow$)}  &  \multicolumn{1}{C{\mycolwidth}}{Rec. ($\uparrow$)} &\multicolumn{1}{C{\mycolwidth}}{F1. ($\uparrow$)} &  \multicolumn{1}{C{\mycolwidthep}}{AUC ($\uparrow$)} & 
 \multicolumn{1}{C{\mycolwidth}}{Ear. ($\uparrow$)} &
 \multicolumn{1}{|C{\mycolwidth}}{Acc. ($\uparrow$)}   & \multicolumn{1}{C{\mycolwidth}}{Rec. ($\uparrow$)} &\multicolumn{1}{C{\mycolwidth}}{F1. ($\uparrow$)} & \multicolumn{1}{C{\mycolwidthep}}{AUC ($\uparrow$)}  & 
 \multicolumn{1}{C{\mycolwidth}}{Ear. ($\uparrow$)} &
\multicolumn{1}{|C{\mycolwidthadd}}{Comp. ($\uparrow$)}\\ 
\midrule
NB~\cite{Watson2001AnES} &  \multicolumn{1}{|C{\mycolwidth}}{54.70} & 23.21 & 32.49 & \underline{59.62} & 18.30 & \multicolumn{1}{|C{\mycolwidth}}{51.30} & 16.07 & 23.93 & 60.10 & 12.16 & \multicolumn{1}{|C{\mycolwidth}}{\underline{53.85}} & 14.28 & 22.53 & \underline{60.47} & 10.49 & \multicolumn{1}{|C{\mycolwidthadd}}{19.98} \\ 
EE~\cite{Lines2015TimeSC} &  \multicolumn{1}{|C{\mycolwidth}}{50.43} & 51.79 & 49.27 & 52.97 & 35.38 & \multicolumn{1}{|C{\mycolwidth}}{52.15}   & 57.14 & 53.30 & 52.92 & 38.05 & \multicolumn{1}{|C{\mycolwidth}}{50.43} & 35.71  & 40.72 & 47.95 & 24.33 & \multicolumn{1}{|C{\mycolwidthadd}}{40.17} \\  \midrule
ECTS~\cite{Xing2012EarlyCO} &  \multicolumn{1}{|C{\mycolwidth}}{52.11} & 30.35 & 37.80 & 54.62 & 1.92 & \multicolumn{1}{|C{\mycolwidth}}{51.24}   & 23.22 & 30.91 & 52.66 & 1.22 & \multicolumn{1}{|C{\mycolwidth}}{51.26} & 17.86 & 25.71 & 48.70 & 2.43 & \multicolumn{1}{|C{\mycolwidthadd}}{16.66} \\ 
MCFEC~\cite{He2015EarlyCO}  &  \multicolumn{1}{|C{\mycolwidth}}{51.29} & 37.50 & 29.58 & 51.01 & 33.93 & \multicolumn{1}{|C{\mycolwidth}}{51.29}  & 33.93 & 28.36 & 47.03 & 30.41 & \multicolumn{1}{|C{\mycolwidth}}{47.03} &  30.35 & 25.37  & 47.43 & 25.86 & \multicolumn{1}{|C{\mycolwidthadd}}{28.91} \\
EARLIEST~\cite{Hartvigsen2019AdaptiveHaltingPN} &  \multicolumn{1}{|C{\mycolwidth}}{\underline{54.74}} & 41.07 & 46.47 & 58.46 & 19.31 & \multicolumn{1}{|C{\mycolwidth}}{53.84} & 25.00 & 33.89 & 54.16 & 16.74 & \multicolumn{1}{|C{\mycolwidth}}{50.41} &  10.71 & 17.16 & 52.71 & 4.80 & \multicolumn{1}{|C{\mycolwidthadd}}{23.06} \\ 
RHC~\cite{hartvigsen2020recurrent} &  \multicolumn{1}{|C{\mycolwidth}}{45.29} & 66.07 & 53.62 & 40.67 & 32.58 & \multicolumn{1}{|C{\mycolwidth}}{48.67} & 66.07 & 53.65 & 44.88 &  40.51 & \multicolumn{1}{|C{\mycolwidth}}{43.68} & 67.85  & 53.37 & 38.00 & \cellcolor[gray]{0.82}{44.03} & \multicolumn{1}{|C{\mycolwidthadd}}{46.29} \\  \midrule
Time2Graph~\cite{cheng2020time2graph} &  \multicolumn{1}{|C{\mycolwidth}}{49.51} & 67.85 &  56.15 & 52.42 & \cellcolor[gray]{0.82}{55.14} & \multicolumn{1}{|C{\mycolwidth}}{\underline{54.60}} & 66.07 & 56.47 & \underline{60.83} & \cellcolor[gray]{0.82}{52.12} & \multicolumn{1}{|C{\mycolwidth}}{52.11} & 66.07 & 56.19  & 54.27 & \cellcolor[gray]{0.82}{50.44} & \multicolumn{1}{|C{\mycolwidthadd}}{\underline{54.41}} \\ 
AliNet~\cite{sun2020knowledge} &  \multicolumn{1}{|C{\mycolwidth}}{49.54} & \underline{87.50} & \underline{62.44} & 57.37 & \cellcolor[gray]{0.82}{42.19} & \multicolumn{1}{|C{\mycolwidth}}{49.53} & \underline{85.71} & \underline{62.00} & 57.20 & \cellcolor[gray]{0.82}{45.76} & \multicolumn{1}{|C{\mycolwidth}}{50.42} &  71.43 & \underline{57.40}  & 54.11 & \cellcolor[gray]{0.82}{38.17} & \multicolumn{1}{|C{\mycolwidthadd}}{51.32} \\ 
KE-GCN~\cite{Yu2021KnowledgeEB} &  \multicolumn{1}{|C{\mycolwidth}}{50.42} & 83.93 &  61.78 & 54.94 & \cellcolor[gray]{0.82}{40.73} & \multicolumn{1}{|C{\mycolwidth}}{52.98} & 80.36 & 61.90 & 54.57 & \cellcolor[gray]{0.82}{41.74} & \multicolumn{1}{|C{\mycolwidth}}{47.89} & \underline{67.86}  &  55.52 & 50.17 & 37.50 & \multicolumn{1}{|C{\mycolwidthadd}}{49.86} \\ \midrule
\emph{\textbf{CAND}} &  \multicolumn{1}{|C{\mycolwidth}}{\textbf{58.11}} & \textbf{92.86} & \textbf{68.39} & \textbf{62.25} & \cellcolor[gray]{0.82}{43.08} & \multicolumn{1}{|C{\mycolwidth}}{\textbf{54.71}}  & \textbf{91.20} & \textbf{65.65} & \textbf{62.67} & \cellcolor[gray]{0.82}{47.32} & \multicolumn{1}{|C{\mycolwidth}}{\textbf{59.85}} & \textbf{73.22}  & \textbf{63.33} & \textbf{61.20} & \cellcolor[gray]{0.82}{46.32} & \multicolumn{1}{|C{\mycolwidthadd}}{\textbf{55.68}} \\
\bottomrule[1.3pt]
\end{tabular}
\label{table:comparison}
\vspace{-0.3em}
\end{table*}

\subsection{Dataset and Experimental Setup}

\subsubsection{\textbf{Dataset and Preprocessing}}
\label{subsubsec:data_preprocess}
We collect a real-world dataset from National Cheng Kung University Hospital (NCKUH)\footnote{The study protocol was approved by the Institutional Review Board (IRB) of NCKUH (No. B-BR-106-044 \& No. A-ER-109-027).}. The dataset includes minute-by-minute vital signs from ICU patients, collected using wearable devices. It also features Apache II scores~\cite{Knaus1985apacheii}, a measure of illness severity, recorded at specific timestamps. Details of the dataset are in Appendix~\ref{appendix:dataset}. We select 99 ICU patients from the dataset with the most vital signs records. Based on ~\cite{Godinjak2016PredictiveVO}, we \textcolor{black}{classify Apache II scores into \textbf{eight levels}: scores 0-4, 5-9, 10-14, 15-19, 20-24, 25-29, 30-34 and above 35, corresponding to different potential mortality rates. In our experiments, an increase in the Apache II level signifies nuanced deterioration, while a decrease indicates recovery. For example, if a patient's Apache II score moves from the 0-4 range (level 1) to 5-9 (level 2), we annotate this patient as deteriorating.} We select heart rate ($\mathcal{X}_1$) and skin temperature ($\mathcal{X}_2$) as the two primary vital signs. 
Based on ~\cite{Rose2015UtilizationOE}, we analyze 8 hours of data prior to each Apache II level change, extracting 175 sets of vital sign measurements (\textcolor{black}{\textbf{84 labeled as deterioration, 91 labeled as recovery}}), each comprising 8-hour time series of heart rates and skin temperatures. Two-thirds of vital sign measurements are used for constructing knowledge structures (Appendix~\ref{appendix:ks_construct}) and training the model, with the remaining third for testing. 
Each domain KS has 4 relation types for different time intervals between concepts. The cross KS includes 5 correlation types, each representing a range of likelihood for simultaneous occurrences of cross-domain concepts (detailed statistics are in Appendix~\ref{appendix:statistic_ks}). To simulate real-world data scarcity, we randomly remove 30\% and 50\% of deteriorating data from the training dataset, creating pruned datasets with pruning levels of \{0\%, 30\%, 50\%\}.

\subsubsection{\textbf{Compared Methods}} Baselines are classified into three categories: feature-based (Naive Bayes (NB)~\cite{Watson2001AnES} and Elastic Ensembles (EE)~\cite{Lines2015TimeSC}), early classification (ECTS~\cite{Xing2012EarlyCO}, MCFEC~\cite{He2015EarlyCO}, EARLIEST~\cite{Hartvigsen2019AdaptiveHaltingPN}, and RHC~\cite{hartvigsen2020recurrent}), and graph embedding  (Time2Graph~\cite{cheng2020time2graph}, KE-GCN~\cite{Yu2021KnowledgeEB} and AliNet~\cite{sun2020knowledge}).
In our setup, Time2Graph constructs a directed graph for a vital sign type from extracted shapelets, training each graph independently. KE-GCN and AliNet, focusing on aligned knowledge graph embeddings, train our two domain KSs concurrently, treating cross-domain concepts as aligned and identical entities. \textcolor{black}{Here, Time2Graph is considered to employ only domain-specific knowledge for training, while AliNet and KE-GCN utilize both domain-specific and cross-domain knowledge. However, unlike our \emph{CAND}, AliNet and KE-GCN do not infer the correlation strengths to influence the learning of each domain KS.} Details of baselines are in Appendix~\ref{appendix:baseline}.

\subsubsection{\textbf{Basic Setup}} Except for early classification methods, to simulate real-time monitoring, other baselines and our \emph{CAND} divide testing time series into 30-minute subsequences, each transmitted sequentially as observed data. Time2Graph, KE-GCN, AliNet, and \emph{CAND} concatenate learned representation vectors to form observed time series representations, used as input for classification with the XGBoost model~\cite{Chen2016XGBoostAS}. The monitoring is halted if nuanced illness deterioration is detected.

\subsubsection{\textbf{Evaluation Metrics.}}
We present results averaged over 3-fold cross-validation on different pruned datasets, covering accuracy \textbf{(Acc.)}, recall \textbf{(Rec.)}, F1-score \textbf{(F1.)}, AUC value \textbf{(AUC)}, and earliness score \textbf{(Ear.)}. The earliness score is defined as $\frac{\text{T}-t}{\text{T}}$, following the setting in ~\cite{gupta2020approaches}. Here, $\text{T}$ is the length (each unit represents one minute) of the complete time series, defined as 480 minutes (8 hours) in Sec.~\ref{subsubsec:data_preprocess}, and $t$ is the length (in minutes) of observed time series for detecting nuanced deterioration. A higher earliness score indicates earlier deterioration detection with fewer observations for vital signs. Additionally, we report overall composite score \textbf{(Comp.)} to summarize average performance across pruned datasets. The composite score is defined as $\frac{\text{F1-score}+\text{earliness score}}{2}$, revealing the model's capability in balancing detection effectiveness with earliness. Note that all evaluation metrics range from 0\% to 100\%, with higher values indicating better performance.

\begin{table}[t]
\small
\centering
\caption{\textcolor{black}{Results (in \%) on the ablation study.}}
\vspace{-1.3em}
\renewcommand{\arraystretch}{0.3}
\setlength{\tabcolsep}{0.6pt}
\begin{tabular}{lC{\mycolwidth}C{\mycolwidth}C{\mycolwidth}C{\mycolwidth}C{\mycolwidth}C{\mycolwidth}C{\mycolwidth}C{\mycolwidth}}
\toprule[1.3pt]
\multirow{2}{*}{Methods} & \multicolumn{4}{|C{\mythreecolwidth}}{$\quad\quad\quad\:$Pruning = 0\%  }  & \multicolumn{4}{|C{\mythreecolwidth}}{$\quad\quad\quad\:$Pruning = 50\%  }  \\ \cmidrule(rl){2-5} \cmidrule(rl){6-9} 
 & \multicolumn{1}{|C{\mycolwidth}}{Acc. ($\uparrow$)}   & \multicolumn{1}{C{\mycolwidth}}{F1. ($\uparrow$)} & 
 \multicolumn{1}{C{\mycolwidthep}}{AUC ($\uparrow$)} &
 \multicolumn{1}{C{\mycolwidth}}{Ear. ($\uparrow$)} &  \multicolumn{1}{|C{\mycolwidth}}{Acc. ($\uparrow$)}  &\multicolumn{1}{C{\mycolwidth}}{F1. ($\uparrow$)} & \multicolumn{1}{C{\mycolwidthep}}{AUC ($\uparrow$)} & \multicolumn{1}{C{\mycolwidth}}{Ear. ($\uparrow$)} \\ \midrule
 \textbf{Full Model} & \multicolumn{1}{|C{\mycolwidth}}{\textbf{58.11}}  &  \textbf{68.39} & \textbf{62.25} & \textbf{43.08} & \multicolumn{1}{|C{\mycolwidth}}{\textbf{59.85}}  &  \textbf{63.33} & \textbf{61.20} & \textbf{46.32}   \\ 
\textbf{w/o conc.} & \multicolumn{1}{|C{\mycolwidth}}{52.11}  & 64.13 & 51.70 & 41.41 & \multicolumn{1}{|C{\mycolwidth}}{55.58}  &  58.67 & 57.86 & 39.40   \\ 
\textbf{w/o corr.} & \multicolumn{1}{|C{\mycolwidth}}{50.42}  & 62.81 & 57.38 & 36.72 & \multicolumn{1}{|C{\mycolwidth}}{56.37} & 59.88 & 58.50 & 43.64  \\ 
\textbf{w/o infer.} & \multicolumn{1}{|C{\mycolwidth}}{51.27} & 63.68 & 58.01 & 40.52 & \multicolumn{1}{|C{\mycolwidth}}{57.29}  & 60.18 & 59.32 & 39.28 \\ 
\bottomrule[1.3pt]
\end{tabular}
\label{table:ablation}
\end{table}

\vspace{-0.5em}
\begin{figure}[h]
\graphicspath{{figs/}}
\subfloat[Pruning=0\%.] {
\label{fig:ab_recall_prune_0}
\includegraphics[width=0.22\textwidth]{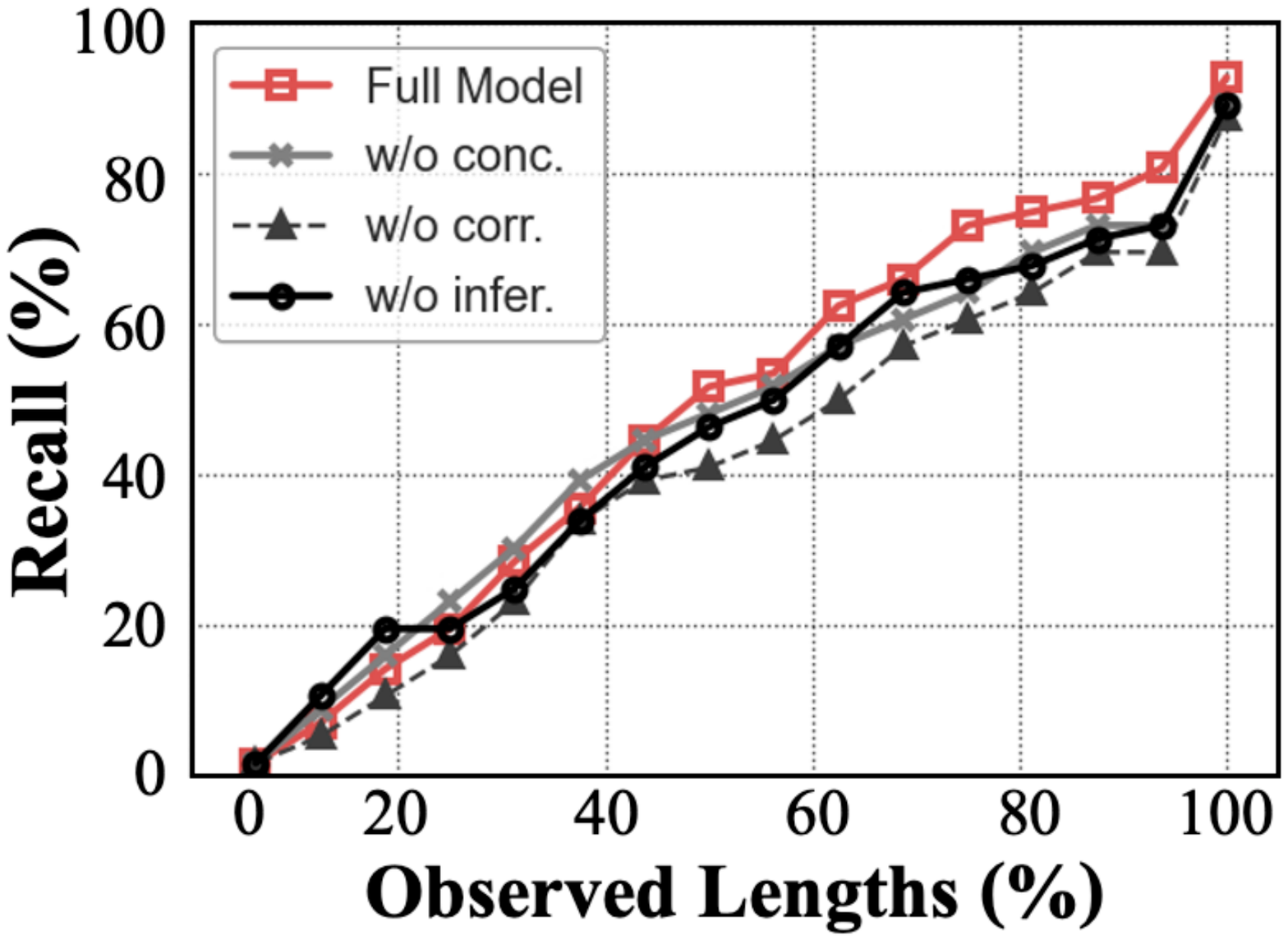}
}
\subfloat[Pruning=50\%.] {
\label{fig:ab_recall_prune_50}
\includegraphics[width=0.22\textwidth]{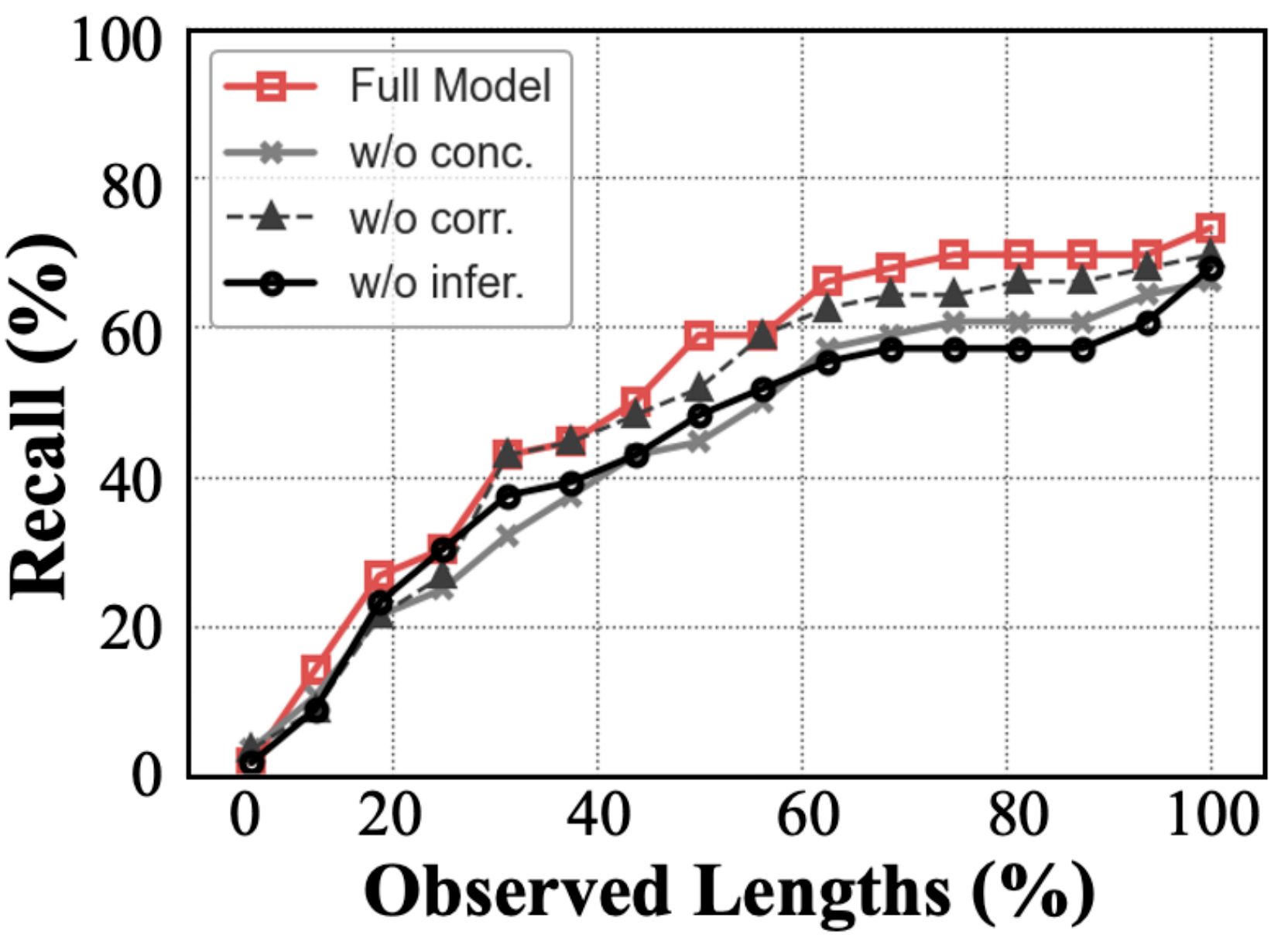}
}
\vspace{-1em}
\caption{\textcolor{black}{Recall performance up to specific observed lengths (\% of total time series) in the ablation study.}}%
\label{fig:ab_recall}%
\end{figure}

\subsection{Comparison Results}
\label{subsec:comparison}
Table~\ref{table:comparison} shows comparison results. Following observations are drawn:

\begin{itemize}[leftmargin=*]
 \item Our \emph{CAND} outperforms baselines in accuracy, recall, F1-score, and AUC across all pruned datasets, which highlights its robustness under different data scarcities. Although the earliness score of \emph{CAND} is lower than Time2Graph, \emph{CAND} significantly exceeds it in other metrics. This indicates that \emph{CAND} greatly enhances detection effectiveness with only a slight increase in observed vital signs. \textcolor{black}{Moreover, \emph{CAND} achieves the highest composite score (55.68\%), which demonstrates its ability to balance detection earliness and effectiveness well.}
 \item Early classification methods (ECTS, MCFEC, EARLIEST, RHC) utilize different halting mechanisms compared to other baselines and our \emph{CAND}. However, their much lower recall and earliness scores indicate limitations in using few vital sign data to detect more deteriorating patients. Graph-based methods show stable performance. This emphasizes the value of domain-specific knowledge, specifically transition relationships among shapelets. Despite this, Time2Graph, which lacks consideration for correlations among vital signs, only reaches about 70\% recall and indicates that around 30\% of undetected deteriorating patients. AliNet and KE-GCN, which treat cross-domain concepts as identical without inferring their correlation strengths, fall short in performance compared to \emph{CAND} across all metrics.
 \item \textcolor{black}{Although the AUC value of our \emph{CAND} is not particularly high from a general perspective (around 62\%), other baselines fail to achieve both high recall and AUC values as effectively as \emph{CAND}. Additionally, our experiments focus solely on detecting deterioration through vital sign monitoring, which can be considered an information-scarce scenario compared to analyzing patients' complete electronic health records. As such, an AUC of 60\% or greater is generally considered desirable~\cite{iyer2016screening}.} 
\end{itemize}

\vspace{-0.1em}
\begin{figure}[h]
\graphicspath{{figs/}}
\begin{center}
\includegraphics[width=0.49\textwidth]{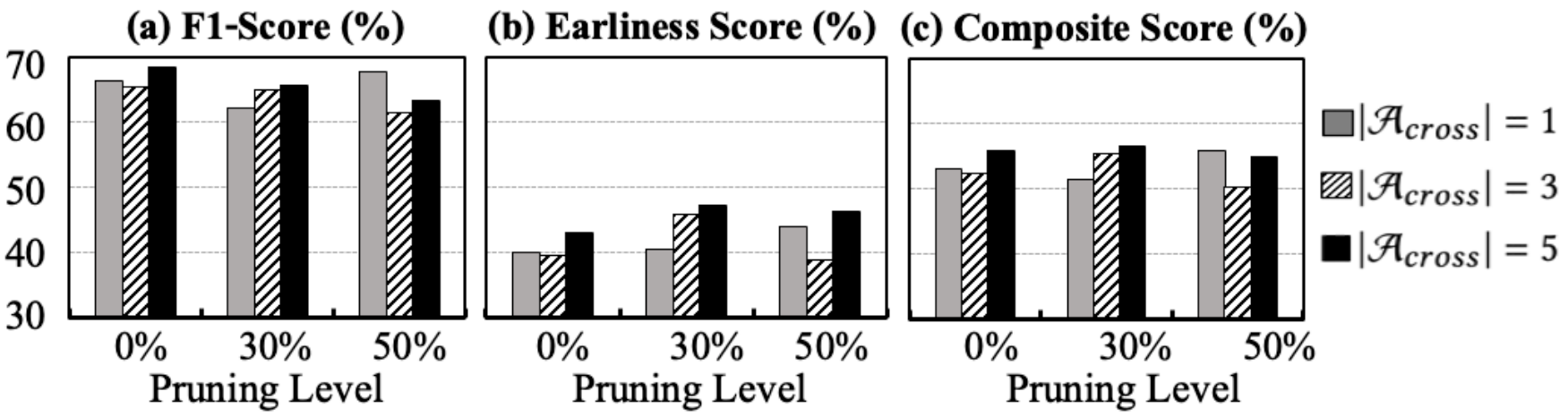}
\end{center}
\vspace{-1.2em}
\caption{\textcolor{black}{Performance of $\emph{CAND}$ with varying  number of ambiguous
correlation types ($|\mathcal{A}_{cross}|$).}}
\label{fig:type_quantity}
\end{figure}

\vspace{-1.2em}
\begin{figure}[h]
\graphicspath{{figs/}}
\subfloat[Performance of varying the weight of contextual loss ($\lambda_c$).] {
\label{fig:param_cont_weight_comp}
\includegraphics[width=0.225\textwidth]{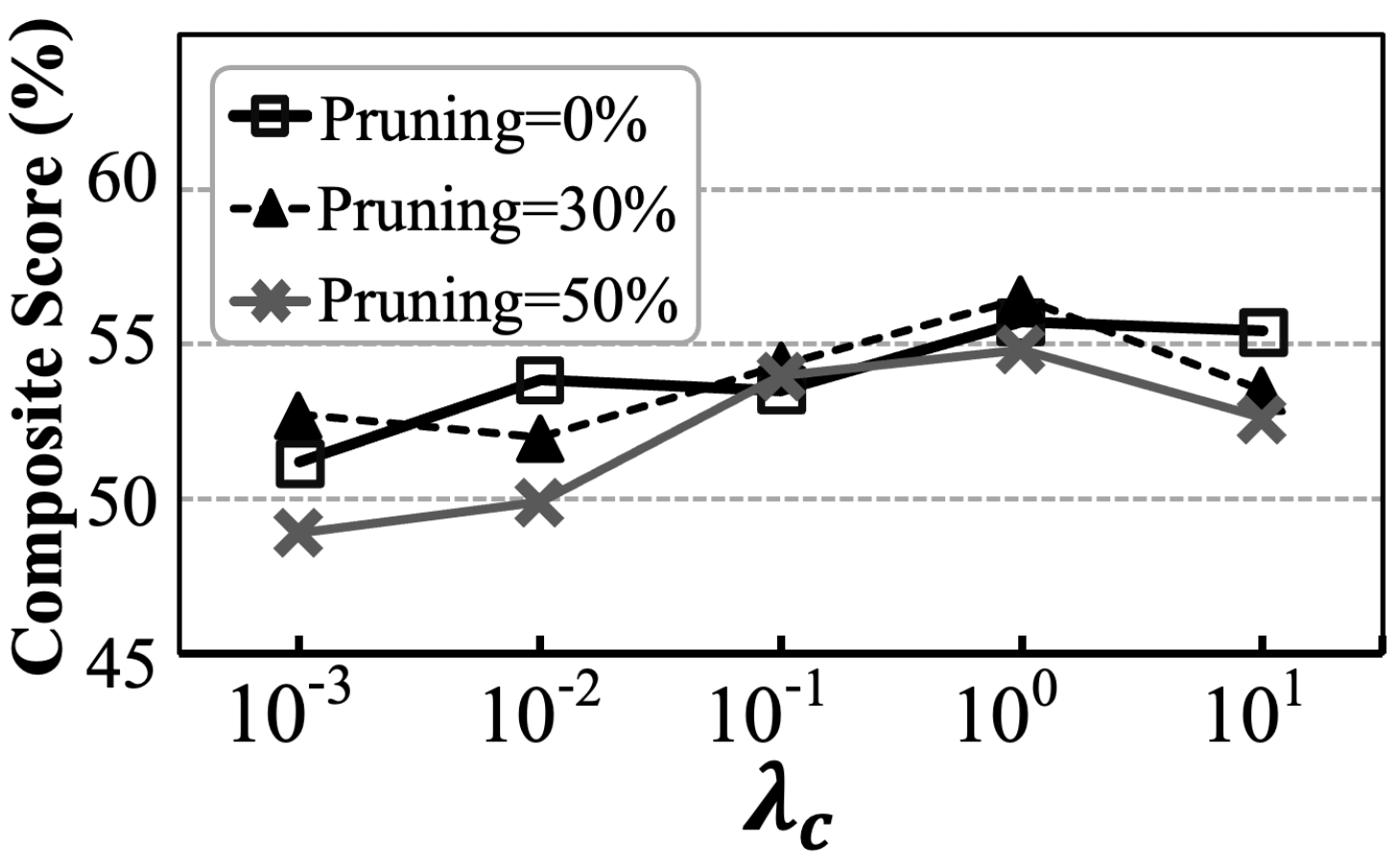}
}
\hspace{0.1mm}
\subfloat[Performance of varying the scale factor ($\xi$).] {
\label{fig:param_scale_comp}
\includegraphics[width=0.22\textwidth]{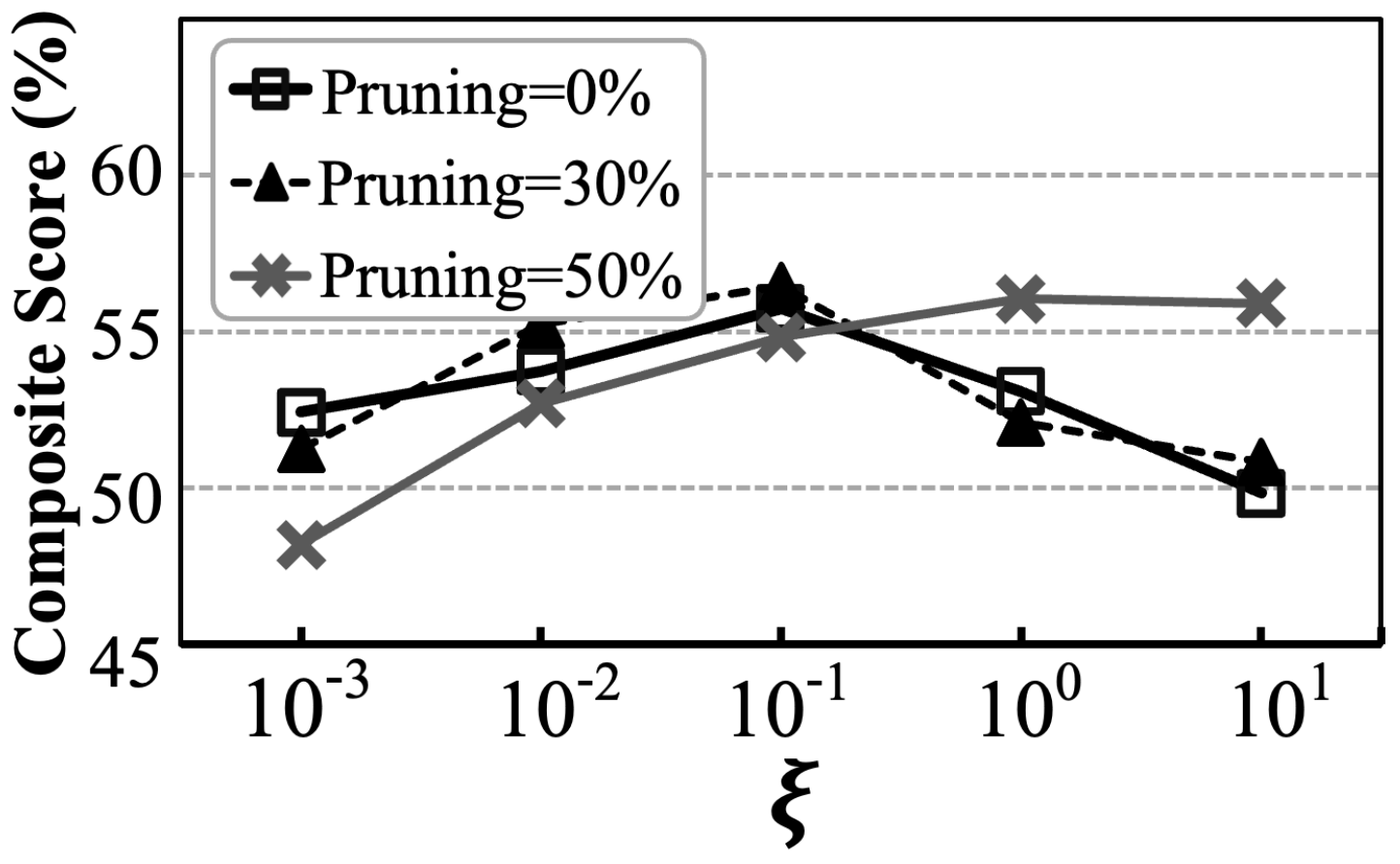}
}
\vspace{-1.2em}
\caption{\textcolor{black}{Composite scores of \emph{CAND} with different settings.}}%
\label{fig:param_comp}%
\end{figure}

\begin{figure*}[t]
\graphicspath{{figs/}}
\begin{center}
\includegraphics[width=0.8\textwidth]{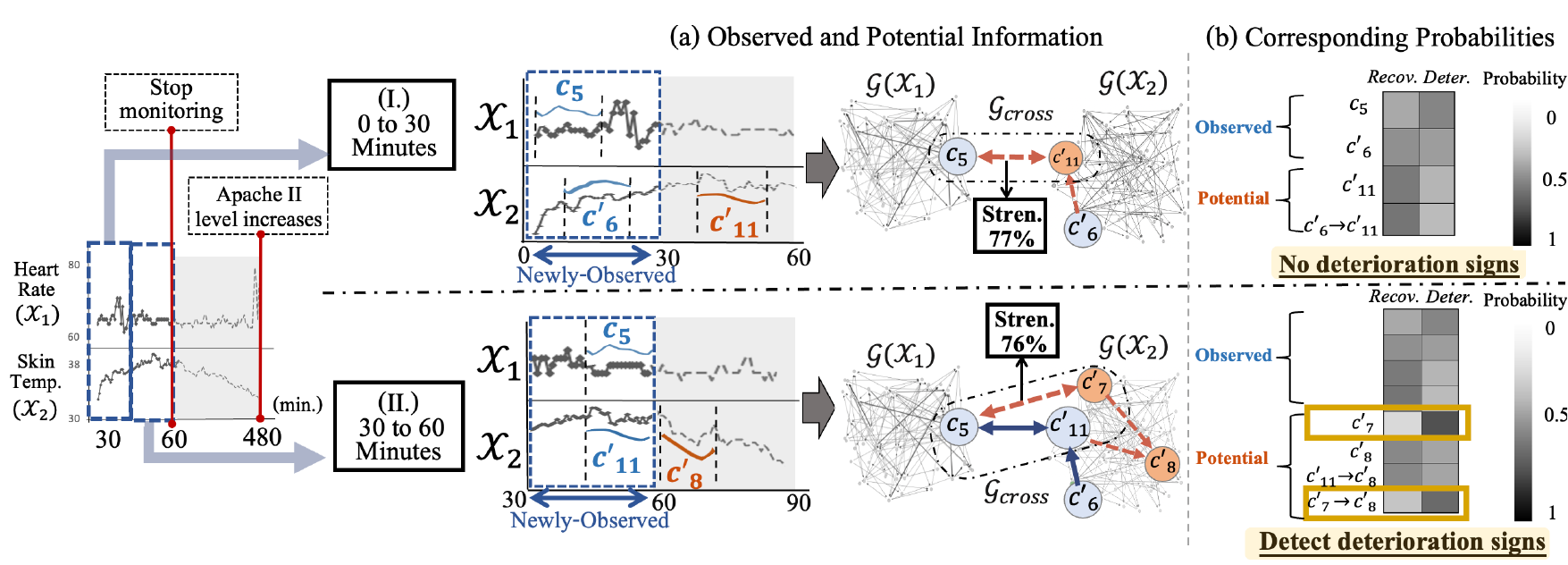}
\end{center}
\vskip -1.3em
\caption{Visualization of the detection process for a deteriorating patient in two detection windows (0 to 30 and 30 to 60 minutes). (a) displays the observed (blue part) and the potential (red part) information identified by \emph{CAND} leveraging inferred correlation strengths (Stren.);(b) depicts corresponding probabilities of observed and potential information in recovering (Recov.) and deteriorating (Deter.) patients from historical data.}
\label{fig:case_study1}
\vspace{-1em}
\end{figure*}

\subsection{Ablation Analysis}

To validate the contributions of main components of \emph{CAND}, we conduct an ablation study by examining the performance after removing concept-view contextual triplets (w/o conc.), correlation-view contextual triplets (w/o corr.), and the inference of correlation strengths (w/o infer.). The results on two pruned datasets (pruning level=\{0\%, 50\%\}) are presented in Table~\ref{table:ablation}. Additionally, Figure~\ref{fig:ab_recall} also shows the recall performance at different observed lengths of vital sign time series. We can find that each component contributes to the overall performance in Table~\ref{table:ablation}. Figure~\ref{fig:ab_recall} shows that Full Model (\emph{CAND}) achieves a faster increase in recall, particularly after 80\% of vital signs are observed. This suggests that while vital signs observed closer to the time of an Apache II level change are inherently more informative, \emph{CAND} is more efficient in capturing these critical signs. The design of \emph{CAND} is verified.

\vspace{-0.2em}
\subsection{Impact of Correlation Type Quantity}

We categorize correlations into 5 types ($|\mathcal{A}_{cross}|=5$) as described in Sec~\ref{subsubsec:data_preprocess}. This categorization can be seen as side information of correlations extracted from data or provided by clinicians. To assess the performance of \emph{CAND} with varying degrees of side information, we conduct experiments with different numbers of correlation types, specifically varying $|\mathcal{A}_{cross}|$ to $\{1,3,5\}$. \textcolor{black}{Figure~\ref{fig:type_quantity} shows that when the number of correlation types is set to 5, performance reaches the highest at 0\% and 30\% pruned datasets. Interestingly, while richer correlation information ($|\mathcal{A}_{cross}|=5$) may marginally improve performance, \emph{CAND} maintains competitive performance even with sparse correlation details ($|\mathcal{A}_{cross}|=1$), showcasing its applicability in scenarios with limited correlation information.}

\vspace{-1em}
\subsection{Parameter Sensitivity Analysis}
\label{subsec:param}

To evaluate the effects of contextual loss weight ($\lambda_c$ in Eq.\ref{eq:loss_cross}) and the scale factor for dynamic margin adjustment ($\xi$ in Eq.\ref{eq:margin_adjust}) on \emph{CAND}, we vary $\lambda_c$ and $\xi$ within $\{10^{-3},10^{-2}, 10^{-1}, 10^{0}, 10^{1} \}$. as shown in Figure~\ref{fig:param_comp}. Figure~\ref{fig:param_cont_weight_comp} reveals that the composite score increases as $\lambda_c$ ranges from $10^{-2}$ to $10^0$ in the 30\% and 50\% pruned datasets, but declines at $10^1$. This highlights the need for balancing weights of different triplets in Eq.\ref{eq:margin_adjust}. Figure~\ref{fig:param_scale_comp} shows that extreme $\xi$ values negatively affect the performance of \emph{CAND} in datasets pruned at 0\% and 30\%, which reveals that too small or large margin adjustments are not optimal. However, in the 50\% pruned dataset, a larger value of $\xi$ leads to better performance. This may indicate that in scenarios with fewer deterioration cases, a more sensitive margin adjustment based on correlation strengths improves detection.

\subsection{Case Study}

We present a case study in Figure~\ref{fig:case_study1}, illustrating the detection process for a patient. This patient was successfully identified as deteriorating after monitoring two vital signs over a 60-minute period (420 minutes before an Apache II level increase). We examine two detection windows: (a)  0 to 30 minutes (undetected window) and (b) 30 to 60 minutes (detected window), focusing on how \emph{CAND} analyzes observed vital signs leveraging inferred correlation strengths.

\noindent\textbf{\textit{Undetected window: 0 to 30 minutes.}} Here, concepts (shapelets) $c_5$ and $c'_6$ match the observed time series of vital signs $\mathcal{X}_1$ and $\mathcal{X}_2$, respectively. Interestingly, both $c_5$ and $c'_6$ suggest the potential occurrence of $c'_{11}$ in the next window, which is not yet observed. The cross-domain concepts pair $(c_5, c'_{11})$ in $\mathcal{G}_{cross}$ shows a high inferred correlation strength (77\%). 
Additionally, $c'_6$ and $c'_{11}$ in domain KS $\mathcal{G}(\mathcal{X}_2)$ show a transition relationship. However, their observed and potential information suggest a low probability of deterioration. Hence, \emph{CAND} does not detect deterioration.

\noindent\textbf{\textit{Detected window: 30 to 60 minutes.}} Concepts $c_{5}$ and $c'_{11}$ are found to match the observed vital signs. Besides $c'_{11}$, $c_5$ also shows a high inferred strength (76\%) with $c'_7$. While there is no direct relationship between $c'_7$ and the observed $c'_{11}$, both show transition relationships with $c'_8$ in domain KS $\mathcal{G}(\mathcal{X}_2)$. Intriguingly, $c'_8$ is found to match the time series of vital sign $\mathcal{X}_2$ in the subsequent time window. Specifically, we discover that $c'_7$ and the transition relationship from $c'_7$ to $c'_8$ in domain KS $\mathcal{G}(\mathcal{X}_2)$ exhibit higher probability in deteriorating patients. This finding suggests that the representations of $c_{5}$ and $c'_{11}$ may contain explicit features of deterioration, which leads \emph{CAND} to detect deterioration signs within this time window. The analysis reveals that the high correlation strength between $c_5$ and $c'_7$ suggests the potential occurrence of $c'_8$. For the raw vital sign data, this indicates that for vital sign $\mathcal{X}_2$, a sharp drop ($c'_8$) following a gentle curve ($c'_{11}$) may be a strong indicator of deterioration.

Regarding these observations, \emph{CAND} shows potential for human-level interpretability in the detection process, \textcolor{black}{providing physicians with clear analytical results to aid in diagnosis.} Also, inferred correlation strengths enhance understanding of vital sign changes, improving detection earliness. \textcolor{black}{Additional experimental results of our \emph{CAND} are provided in Appendix~\ref{appendix:exp}.}

\vspace{-1em}

\section{Conclusion}

We develop \emph{CAND}, a framework that utilizes domain-specific and cross-domain knowledge of vital signs for early detecting nuanced illness deterioration. \emph{CAND} employs multi-view context augmentation and Bayes-based inference to model correlations with ambiguous strengths, guiding joint modeling of multiple knowledge structures. Applied to the real ICU dataset, \emph{CAND} effectively balances detection earliness and effectiveness, showing potential for human-level interpretability. Additionally, \emph{CAND} leverages low-cost vital sign data, offering potential for remote healthcare development.

\bibliographystyle{ACM-Reference-Format}
\bibliography{bibliography}

\appendix

\section{Notations}
\label{appendix:notation}

In this section, we outline critical notations and their corresponding descriptions used throughout this paper to enhance clarity.

\begin{table}[h]
\small
\centering
\caption{Notations.}
\vspace{-1.2em}
\renewcommand{\arraystretch}{1.2}
\begin{tabular}{c|c}
\toprule[1.3pt]
\textbf{Notations} & \textbf{Descriptions} \\ \hline\hline
$\mathcal{X}$ & a specific type of vital sign \\
$\mathcal{D}_{\mathcal{X}}$ & a historical dataset for vital signs$\mathcal{X}$\\
$T_{\mathcal{X}}\in \mathcal{D}_{\mathcal{X}}$ & a time series for vital sign $\mathcal{X}$\\
$\mathcal{G}(\mathcal{X})$ & the domain KS built from $D_{\mathcal{X}}$\\
$\mathcal{C}_{\mathcal{X}}$, $\mathcal{T}_{\mathcal{X}}$ & the concept set and the relation set of  $\mathcal{G}(\mathcal{X})$\\
$(c_i, \tau, c_j) \in \mathcal{G}(\mathcal{X})$ & a triplet in $\mathcal{G}(\mathcal{X})$, $c_i, c_j\in \mathcal{C}_{\mathcal{X}}$, $\tau \in \mathcal{T}_{\mathcal{X}}$\\
$\mathcal{G}_{cross}$ & a cross-domain KS\\
$\mathcal{C}_{cross}$, $\mathcal{A}_{cross}$ & the concept set and the correlation set of $\mathcal{G}_{cross}$ \\
$(c, a, c') \in \mathcal{G}_{cross}$ & a triplet in $\mathcal{G}_{cross}$, $c, c
\in \mathcal{C}_{cross}$, $a \in \mathcal{A}_{cross}$\\
$(c_{cont}, a, c'_{cont})$ & concept-view contextual triplet of $(c, a, c')$ \\
$(c, a_{cont}, c')$ & correlation-view contextual triplet of $(c, a, c')$ \\
$\omega\in \Omega$ & an ambiguity type\\
$(c,c')\in \mathcal{U}$ & a pair of cross-domain concepts\\
$\mathcal{E}_{c,c',\omega}$ & the correlation strength of $\omega$ with $(c,c')$ \\
$\gamma_{\nu}$ & the dynamic margin for cross KS loss\\
$\xi$ & the scale factor to control the value of $\gamma_{\nu}$\\
$\lambda_c$ & the weight of contextual loss in cross KS loss\\
\bottomrule[1.3pt]
\end{tabular}
\renewcommand{\arraystretch}{1}
\label{table:notation}
\end{table}

\section{Dataset Description}
\label{appendix:dataset}

This study uses wearable devices to monitor patients in real-time, gathering vital sign data from ICU patients at National Cheng Kung University Hospital (NCKUH).

\noindent\textbf{Patient Criteria.} The study involves ICU patients at NCKUH between August 2020 and December 2021. \textcolor{black}{The following conditions lead to the exclusion of certain patients: (i) Those who are physically or mentally disabled and unable to collaborate with researchers, (ii) Patients who may experience fever due to factors like rheumatic immunity, allergies, blood transfusions, malignant tumors, or other diseases, and (iii) Patients suffering from severe peripheral vascular disease.}

\noindent\textbf{Data Collection.} Patients are required to wear devices to track heart rates and skin temperatures minutely. Additionally, patient data are collected through the hospital electronic system, including:

\begin{itemize}[leftmargin=*]
    \item Basic Information: \textcolor{black}{age, occupation, medication usage, gender, BMI, and disease diagnosis}.
    \item Vital Signs: body temperature, respiration rate, heart rate, blood pressure, and blood oxygen levels.
    \item \textcolor{black}{Severity of Illness (Apache II scores~\cite{Knaus1985apacheii}): the Apache II score (Acute Physiology and Chronic Health Evaluation II) is a widely used evaluation tool to assess the severity of disease in ICU patients, which can be treated as a useful indicator of illness deterioration or recovery over time~\cite{naved2011apache}. Generally, a higher Apache II score corresponds to a more severe disease, and it is derived from a patient's age, medical history, and current physiological measurements, such as respiratory rate, serum sodium, blood pH, white blood cell count, Glasgow Coma Scale, among others. In our dataset, each patient has at least two Apache II score records, taken at different times during their ICU stay.}
\end{itemize}

Each patient is required to wear the device for at least 14 days. This paper primarily focuses on data from wearable devices (heart rates, skin temperatures) and Apache II scores, with statistics for 99 selected ICU patients detailed in Table~\ref{table:dataset}.

\vspace{1em}
\begin{table}[!ht]
\small
\centering
\caption{Statistics (Statis.) of characteristic (Charac.) for selected 99 ICU patients.}
\vspace{-1.2em}
\renewcommand{\arraystretch}{0.8}
\setlength{\tabcolsep}{1.6pt}
\begin{tabular}{c|ccccccccc}
\toprule[1.3pt]
 & \multicolumn{2}{c|}{\textbf{Gender}} & \multicolumn{5}{c|}{\textbf{Age}} & \multicolumn{2}{c}{\textbf{Death}} \\ \cmidrule(rl){2-3}\cmidrule(rl){4-8}\cmidrule(rl){9-10}
\multicolumn{1}{c|}{\textbf{Charac.}} & Female & \multicolumn{1}{c|}{Male} & 0-20 & 21-40 & 41-60 & 61-80 & \multicolumn{1}{c|}{81-100} & Yes & No \\ \midrule
\multicolumn{1}{c|}{\textbf{Statis.} (\%)} & 32.4 & \multicolumn{1}{c|}{67.6} & 2.0 & 8.1 & 23.2 & 56.6 & \multicolumn{1}{c|}{10.1} & 3.0 & 97.0 \\ \midrule[1pt]
\multicolumn{1}{c|}{\textbf{Charac.}} & \multicolumn{4}{c|}{\thead{\textbf{Avg. stay in ICUs} \\ \textbf{/ patient} (days)}} & \multicolumn{5}{|c} {\thead{\textbf{Avg. value of the Apache II} \\ \textbf{ score  (std.)}}}\\
\midrule
\multicolumn{1}{c|}{\textbf{Statis.} }& \multicolumn{4}{c|}{29.63}  & \multicolumn{5}{|c} {16.39 (6.4)} \\ 
\bottomrule[1.3pt]
\end{tabular}
\label{table:dataset}
\end{table}

\vspace{1em}
\section{Reproducibility} 
\subsection{Implementation Details of \emph{CAND}} 
\label{appendix:implement_detail}

\subsubsection{\textbf{Hyper-parameters.}}
For the default settings, we set the embedding dimension to 256, the shapelet length to 15, and the number of shapelets for heart rate and skin temperature data ($\mathcal{D}_{\mathcal{X}_1}$ and $\mathcal{D}_{\mathcal{X}_2}$) to 60 and 80, respectively. The lengths for explorations ($L$ in Sec \ref{subsubsec:domain_view}) and the maximum lengths of correlation paths ($L'$ in Sec. \ref{subsubsec:concept_view}) are both set to 7. The number of the explorations is set to 20 ($N$ in Eq.~\ref{eq:domain_triplet_formulation}), and the parameter $\varphi$ in Eq.~\ref{eq:traversal_prob} is set to 4. The scale factor $\xi$ in Eq.~\ref{eq:margin_adjust} is set 0.1. The default values of $\lambda_c$ in Eq.~\ref{eq:loss_cross} is 1, and $\epsilon$ in Eq.~\ref{eq:lost_mila} is set to 0.5.

\subsubsection{\textbf{Knowledge Structure Construction.}} 
\label{appendix:ks_construct}
\leavevmode\newline
\textbf{\underline{Domain KS.}}
An example of constructing domain KS $\mathcal{G}(\mathcal{X}_1)$ is shown in Figure~\ref{fig:domain_ks_construct}. We first extract concepts (shapelets) from historical dataset $\mathcal{D}_{\mathcal{X}_1}=\bigl \{ T_{\mathcal{X}_1}^i \bigr \}_{i=1}^N$ (Def.1) using the existing method~\cite{Grabocka2014LearningTS}. Setting the shapelet length ($L$) to 15 (15 minutes) and the number of shapelets ($K$) to 60, we can obtain a set of concepts (shapelets) $\mathcal{C}_{\mathcal{X}_1}\in \mathbb{R}^{K\times L}$, as shown in Figure~\ref{fig:domain_ks_construct} (a)(b). Subsequently, for each time series $T_{\mathcal{X}_1}^i\in \mathcal{D}_{\mathcal{X}_1}$, we estimate matching scores to determine which concepts in $\mathcal{C}_{\mathcal{X}_1}$ are matched to $T_{\mathcal{X}_1}^i$ (as in Figure~\ref{fig:domain_ks_construct}(c)).
Firstly, we divide $T_{\mathcal{X}_1}^i$ into $n$ subsequences of length $L$, represented as $T_{\mathcal{X}_1}^i=\{v_1, ..., v_n\}$, with each subsequence $v_k=\{x_{l*(k-1)+1},...,x_{l*(k-1)+L}\}$, where each $x\in \mathbb{R}$ is a real-number reading of vital sign $\mathcal{X}_1$. The distance $\hat{d}(c,v)$ between a subsequence $v\in T_{\mathcal{X}_1}^i$ and a concept $c\in \mathcal{C}_{\mathcal{X}_1}$ is calculated as the average of their Euclidean and Dynamic Time Warping (DTW) distances. Therefore, the matching score $\delta_c$ of concept $c$ to subsequence $v$ is designed as follows:

\vspace{0.5em}
\begin{figure}[h]
\graphicspath{{figs/}}
\begin{center}
\includegraphics[width=0.45\textwidth]{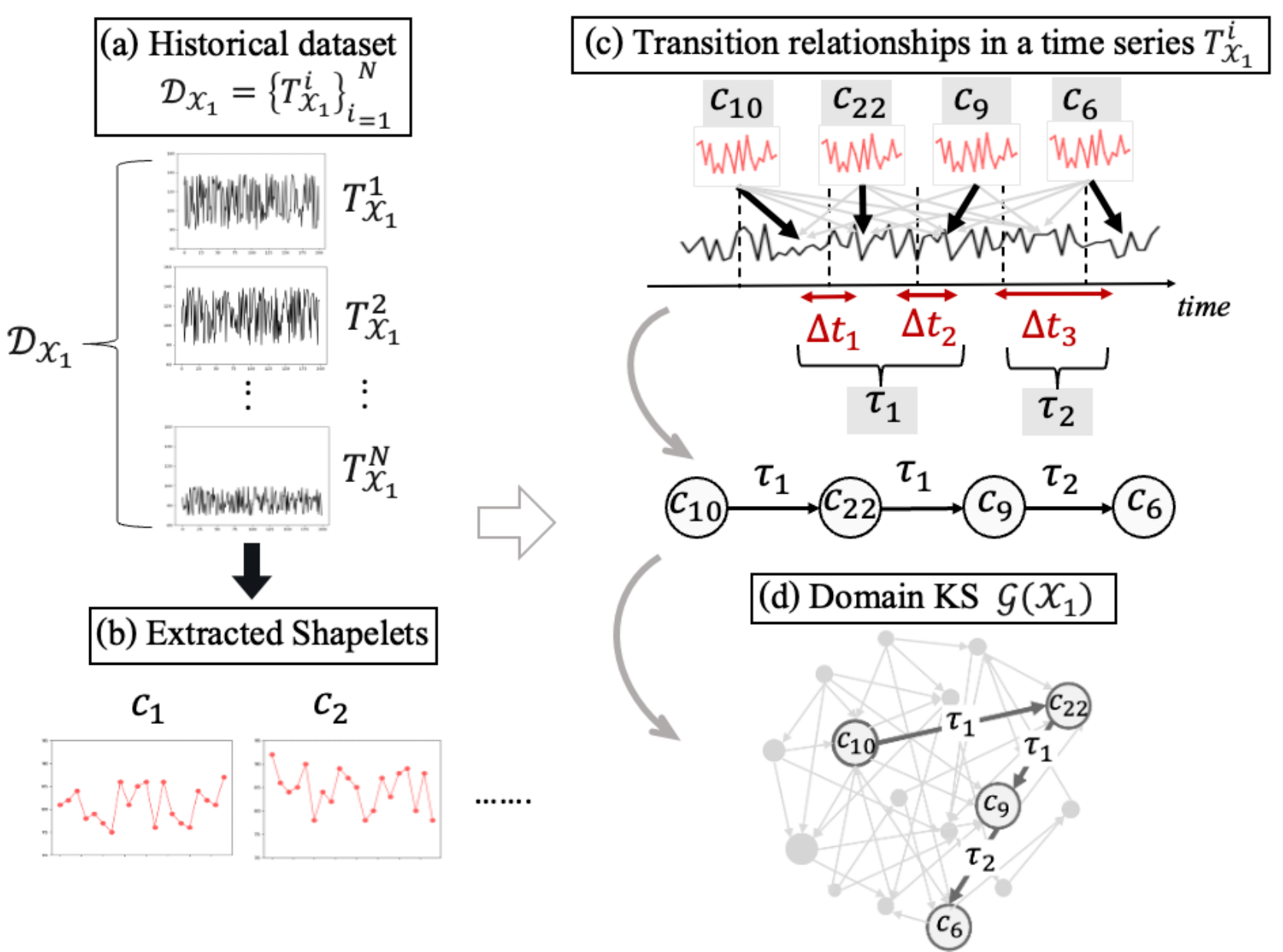}
\end{center}
\vskip -1em
\caption{An example of constructing domain KS $\mathcal{G}(\mathcal{X}_1)$.}
\label{fig:domain_ks_construct}
\end{figure}

\vspace{-1em}
\begin{equation}
\label{eq:alloc_score}
\delta_c=\frac{\text{max}_{c'\in \mathcal{C}_{\mathcal{X}_1}, v'\in T_{\mathcal{X}_1}^i}\hat{d}(c', v')-\hat{d}(c, v)}{\text{max}_{c'\in \mathcal{C}_{\mathcal{X}_1}, v'\in T_{\mathcal{X}_1}^i}\hat{d}(c', v')-\text{min}_{c'\in \mathcal{C}_{\mathcal{X}_1}, v'\in T_{\mathcal{X}_1}^i}\hat{d}(c', v')}.
\end{equation}

The concept $c$ is assigned to subsequence $v$ if  $\delta_c\geq 0.7$ and  $c$ is the top 3 similiest concepts to $v$ among all concepts in $\mathcal{C}_{\mathcal{X}_1}$.

Next, we define the relation set $\mathcal{T}_{\mathcal{X}_1}$, with each relation $\tau\in \mathcal{T}_{\mathcal{X}_1}$ representing a specific time interval range between two concepts in each time series $T_{\mathcal{X}_1}^i\in \mathcal{D}_{\mathcal{X}_1}$. Specifically, we categorize these intervals into four ranges: $\tau_1$ (0-30 minutes), $\tau_2$ (30-60 minutes), $\tau_3$ (60-90 minutes), and $\tau_4$ (up to 90 minutes). Each $\tau_i \in \mathcal{T}_{\mathcal{X}_1}$. A triplet $(c_i, \tau, c_j)$ is formed if both $c_i$ and $c_j$ from $\mathcal{C}_{\mathcal{X}_1}$ are assigned to the same time series, with $c_i$ occurring before $c_j$ within the interval $\tau$. As revealed in Figure~\ref{fig:domain_ks_construct} (d), this forms domain KS $\mathcal{G}(\mathcal{X}_1)=\{(c_i, \tau, c_j)|c_i, c_j \in \mathcal{C}_{\mathcal{X}_1}, \tau \in \mathcal{T}_{\mathcal{X}_1}\}$.  A similar approach is used to construct domain KS $\mathcal{G}(\mathcal{X}_2)$.

\vspace{0.5em}
\noindent\textbf{\underline{Cross KS.}} Upon constructing the domain KSs $\mathcal{G}(\mathcal{X}_1)$ and $\mathcal{G}(\mathcal{X}_2)$, we estimate the likelihood of simultaneous occurrences for each concept pair $(c,c')$, where $c\in \mathcal{G}(\mathcal{X}_1)$ and $c'\in \mathcal{G}(\mathcal{X}_2)$. These likelihoods (excluding zero values) are divided into five ranges of equal intervals, forming the set of ambiguous correlations $\mathcal{A}_{cross} = \{a_1, a_2, ..., a_5\}$. Consequently, cross KS can be formed as $\mathcal{G}_{cross} = \{(c, a, c')|c, c'\in \mathcal{C}_{cross}, a \in \mathcal{A}_{cross}\}$, where $\mathcal{C}_{cross} \subset \mathcal{C}_{\mathcal{X}_1} \cup \mathcal{C}_{\mathcal{X}_2}$.

\vspace{1em}
\begin{table}[h]
\small
\centering
\caption{Statistics of knowledge structures constructed in different pruned datasets.}
\vspace{-1em}
\renewcommand{\arraystretch}{1}
\setlength{\tabcolsep}{2pt}
\begin{tabular}{lc|ccc}
\toprule[1.3pt]
\multicolumn{2}{c}{Datasets} & \multicolumn{1}{|c}\#Concepts & \makecell[c]{\#Relations or \\ \#Correlations} & \#Triplets \\ \midrule
\multirow{3}{*}{Pruning=0\%} & $\mathcal{G}(\mathcal{X}_1)$ & 56 & 4  & 1441 \\
 & $\mathcal{G}(\mathcal{X}_2)$ & 79 & 4 & 4085 \\
 & $\mathcal{G}_{cross}$ & 39 & 5 & 98 \\ \midrule
\multirow{3}{*}{Pruning=30\%} & $\mathcal{G}(\mathcal{X}_1)$ & 54 & 4 & 1486 \\
 & $\mathcal{G}(\mathcal{X}_2)$ & 78 & 4 & 4277 \\
 & $\mathcal{G}_{cross}$ & 44 & 5 & 150 \\ \midrule
\multirow{3}{*}{Pruning=50\%} & $\mathcal{G}(\mathcal{X}_1)$ & 56 & 4 & 1272 \\
 & $\mathcal{G}(\mathcal{X}_2)$ & 80 & 4 & 4755 \\
 & $\mathcal{G}_{cross}$ & 47 & 5 & 176\\
\bottomrule[1.3pt]
\end{tabular}
\label{table:ks_statistic}
\end{table}
\vspace{0.5em}

\subsubsection{\textbf{Statistics of Knowledge Structures in \emph{CAND}}}
\label{appendix:statistic_ks}
In Table~\ref{table:ks_statistic}, we provide summary statistics for the knowledge structure constructed for one of the folds in our 3-fold cross-validation. The default numbers of shapelets (concepts) extracted from datasets $\mathcal{D}_{\mathcal{X}_1}$ and $\mathcal{D}_{\mathcal{X}_2}$ are set to 60 and 80, respectively. For the 30\% and 50\% pruned datasets, during preprocessing, any extracted shapelet that does not match the time series in these datasets is excluded from being a concept in the knowledge structure.

\subsubsection{\textbf{Representation Formulation for Time Series}} 
\label{appendix:represent_timeseries}
After finishing the training of domain KSs and the cross KS.
For the $i^{th}$ set of vital sign time series $\{T^i_{\mathcal{X}_1}, T^i_{\mathcal{X}_2}\}$ in training data, where $T^i_{\mathcal{X}_1}\in \mathcal{D}_{\mathcal{X}_1}$ and  $T^i_{\mathcal{X}_2}\in \mathcal{D}_{\mathcal{X}_2}$, we formulate the representation of the $i^{th}$ measurement based on the occurred triplets in $T^i_{\mathcal{X}_1}$ and $T^i_{\mathcal{X}_2}$. Let $V(T^i_{\mathcal{X}_1})$  be triplet sets including triplets occurred in $T^i_{\mathcal{X}_1}$, where each triplet $\nu_n \in V(T^i_{\mathcal{X}_1})$ represents the $n^{th}$ occurred triplet in $T^i_{\mathcal{X}_1}$. The important of triplet $\nu_n$ is designed as:

\begin{equation}
\label{eq:triplet_importance}
 \mu(\nu_n)=\rho^{|V(T^i_{\mathcal{X}_1})|-n}
\end{equation}
where $\rho$, set at 0.8, assigns higher importance to triplets that occur later. Then, the representation of $T^i_{\mathcal{X}_1}$ can be formulated as follows:

\begin{equation}
\label{eq:singletime_represent}
\bold{\Psi}_{\mathcal{X}_1}^i=\sum_{\nu_n=(c_i,\tau,c_j)\in V(T^i_{\mathcal{X}_1})}^{} \frac{\mu(\nu_n) \cdot \Bigl[\mathbf{c}_i||\bm{\tau}||\mathbf{c}_{j}\Bigr]}{|V(T^i_{\mathcal{X}_1})|},
\end{equation}
where $\mathbf{c}_i$, $\bm{\tau}$ and $\mathbf{c}_{j}$ are representation vectors of concept $c_i$, relation $\tau$ and concept $c_j$, respectively.

The representation $\bold{\Psi}_{\mathcal{X}_2}^i$ of vital sign time series $T^i_{\mathcal{X}_2}$ is formulated similarly. Therefore, the final representation of the $i^{th}$ set of vital sign time series $\{T^i_{\mathcal{X}_1}, T^i_{\mathcal{X}_2}\}$  is formed by concatenating $\bold{\Psi}_{\mathcal{X}_1}^i$ and $\bold{\Psi}_{\mathcal{X}_2}^i$ as follows:

\vspace{-1em}
\begin{equation}
\label{eq:fulltimeseries_represent}
\bold{\Psi}_{\text{vital signs}}^i=[\bold{\Psi}_{\mathcal{X}_1}^i||\bold{\Psi}_{\mathcal{X}_2}^i].
\end{equation}

In the testing phase, representations for newly observed vital signs are constructed similarly.

\subsection{\textcolor{black}{Supplementary Descriptions of the Guided Exploration.}} 
\label{appendix:detail_guided}

\begin{figure}[h]
\graphicspath{{figs/}}
\begin{center}
\includegraphics[width=0.48\textwidth]{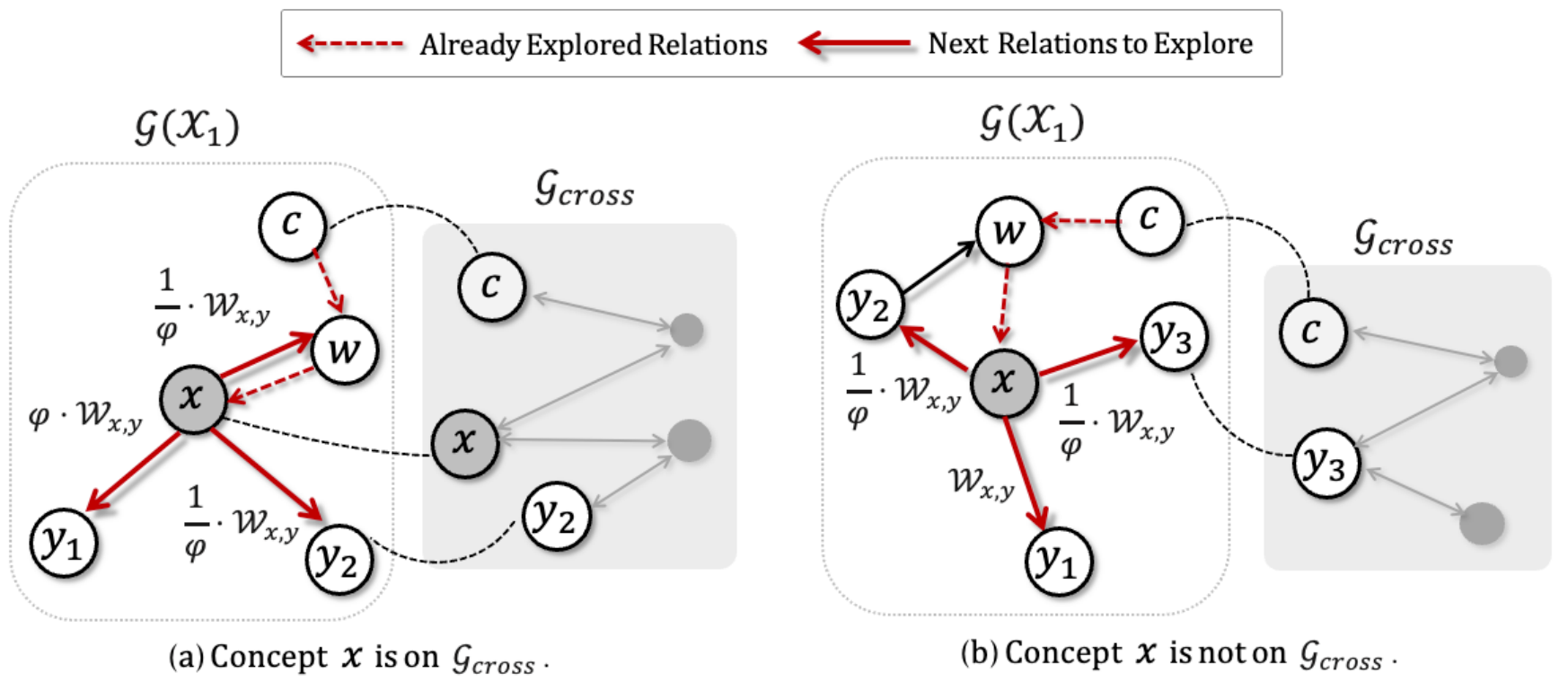}
\end{center}
\vskip -1.2em
\caption{The example of the value of the unnormalized probability $\gamma^{x\to y}$ in the guided exploration.}
\label{fig:detail_exploration}
\end{figure}

In Sec~\ref{subsubsec:domain_view}, we employ a guided exploration, inspired by node2vec~\cite{grover2016node2vec}, to formulate \textit{concept-view} contextual triplets. To further how describe the designed unnormalized probability (Eq.~\ref{eq:traversal_prob}) controls the guided exploration, we present two scenarios in Figure~\ref{fig:detail_exploration}: when the current concept $x$ is on $\mathcal{G}_{cross}$ (Figure 1(a)) and when it's not on $\mathcal{G}_{cross}$ (Figure~\ref{fig:detail_exploration}b), by setting $\varphi > 1$ in Eq.~\ref{eq:traversal_prob}.

For Figure~\ref{fig:detail_exploration}a,  $\gamma^{x\rightarrow y}$ is assigned a smaller value ($\frac{1}{\varphi}  \cdot \mathcal{W}_{x,y}$) if: 
\begin{enumerate}[leftmargin=*]
    \item The previous visited concept $w$ is revisited (e.g., $w$ in Figure~\ref{fig:detail_exploration}a) or the next concept $y$ is one step away from $w$ ($d^{w,y}\leq 1$).
    \item The next concept $y$ belongs to $\mathcal{G}_{cross}$ (e.g., $y_2$ in ~\ref{fig:detail_exploration}a).
\end{enumerate}

When $d^{w,y}= 2$ and $x\in \mathcal{G}_{cross} ,\: y\notin \mathcal{G}_{cross}$ (e.g., $y_1$ in Figure~\ref{fig:detail_exploration}a), $\gamma^{x\rightarrow y}$ is assigned a larger value ($\varphi \cdot \mathcal{W}_{x,y}$). Similarly, in Figure~\ref{fig:detail_exploration}b, $\gamma^{x\rightarrow y}$ is given a lower value ($\frac{1}{\varphi}  \cdot \mathcal{W}_{x,y}$) either if $d^{w,y}\leq 1$ (e.g.,$y_2$ in Figure~\ref{fig:detail_exploration}b) or if $y\in \mathcal{G}_{cross}$ (e.g.,$y_3$ in Figure~\ref{fig:detail_exploration}b).

This approach enables a depth-first search that focuses on exploring contexts most likely to be influenced by concept $c$ deep within its domain KS ($\mathcal{G}(\mathcal{X}_1)$), while avoiding revisits.

\subsection{Baseline Settings.} 
\label{appendix:baseline}

We compare \emph{CAND} with baselines in the following four categories. For feature-based methods, we select \textbf{Naive Bayes}~\cite{Watson2001AnES} and Elastic Ensembles (\textbf{EE})~\cite{Lines2015TimeSC}, using statistical features (mean, skew, standard deviation, kurtosis) from 30-minute intervals of patients' vital sign time series as input.

For early classification methods, we choose:

\begin{itemize}[leftmargin=*]
    \item \textbf{ECTS}~\cite{Xing2012EarlyCO}: ECTS is a prefix-based method aiming for early prediction on univariate time series. In our scenario, ECTS is applied individually to each vital sign time series. A patient is considered as deteriorating only when ECTS detects deterioration signs in time series from all types of vital signs. The earliest timestep at which deterioration is detected in any vital sign is then utilized for estimating the earliness score.
    \item \textbf{MCFEC}~\cite{He2015EarlyCO}:  MCFEC is a shapelet-based method for the early classification of multivariate time series, utilizing distinctive and high-quality shapelets as core features. We employ its MCFEC-QBC (query by committee) classifier, which classifies time series based on the majority class among all matched shapelets.
    \item \textbf{EARLIEST}~\cite{Hartvigsen2019AdaptiveHaltingPN}: EARLIEST is a reinforcement learning-based method that supports early classification of both univariate and multivariate time series. It employs an RNN to model the transition dynamics of time series in conjunction with a policy network that decides at each timestep whether or not to halt the RNN and generate a prediction.
    \item \textbf{RHC}~\cite{hartvigsen2020recurrent}: RHC extends EARLIEST~\cite{Hartvigsen2019AdaptiveHaltingPN} to support early multi-label classification for multivariate time series. It employs a transition model that jointly represents multivariate time series data and the conditional dependencies between labels. This approach potentially captures richer information, benefiting even single-label time series classification in our scenario.
\end{itemize}

\textcolor{black}{For graph embedding methods, Time2Graph inherently incorporates shapelet transitions into a graph structure and is considered to employ only domain-specific knowledge for training. KE-GCN and AliNet, both aligned knowledge graph embedding models, simultaneously train our designed domain KSs, and the cross-domain concepts in our cross KS are regarded as aligned entities in KE-GCN and AliNet. Therefore, KE-GCN and AliNet are considered to utilize both domain-specific and cross-domain knowledge.} The obtained embeddings are used to formulate time series representations in a similar way as our \emph{CAND}. The following are detailed descriptions.

\begin{itemize}[leftmargin=*]
    \item \textbf{Time2Graph}~\cite{cheng2020time2graph}: Time2Graph learns \textit{time-aware shapelets} for time series representations. A set of time series is converted into a directed graph, where nodes are shapelets and edges represent transitions between shapelets. DeepWalk~\cite{perozzi2014deepwalk} is used to derive shapelet representations, which are utilized to form the time series representation by considering the distance between shapelets and the time series. In our experiments, Time2Graph is employed and trained separately for the historical dataset of each vital sign. Representations of each vital sign time series are concatenated for classification as our method (Eq.~\ref{eq:fulltimeseries_represent}). Note that Time2Graph only considers occurrence orders of shapelets, ignoring time intervals between them, and concentrates on the learning for a single graph.
    \item \textbf{AliNet}~\cite{sun2020knowledge}: AliNet is an embedding model for entity-aligned knowledge graphs (KGs), uses an attention mechanism in GNNs (Graph Neural Networks) for multi-hop neighborhood aggregation to minimize representation differences of aligned entities across different KGs. In our experiments, AliNet trains our designed domain KSs simultaneously, treating cross-domain concepts in the cross KS as aligned entities.
    \item \textbf{KE-GCN}~\cite{Yu2021KnowledgeEB}: KE-GCN is an embedding model for entity-aligned or relation-aligned knowledge graphs (KGs), which combines GCNs in graph-based belief propagation and the knowledge graph embedding methods. In our experiments, KE-GCN is applied with the same settings as AliNet, training domain-specific knowledge structures (KSs) and treating cross-domain concepts as aligned entities.
\end{itemize}

\section{Additional Results.} 
\label{appendix:exp}
\subsection{Training Time} 
\label{appendix:time}

In this subsection, we evaluate the computational cost of our \emph{CAND} framework compared to baselines. Experiments are performed on a system equipped with 12 CPU cores and 64GB RAM, running on CUDA version 12.1. We evaluate the average training time over 10 runs for each method, excluding preprocessing time, and presented the results in Table~\ref{table:training_time}. Our findings indicate that the training time of most methods decreases as the pruning level of the dataset increases. While \emph{CAND} requires more training time than most methods, excluding MCFEC and Time2Graph, our \emph{CAND} achieves a more accurate inference of correlation strengths among graphs, leading to significantly improved detection effectiveness and earliness. Note that the training time of \emph{CAND} will not increase linearly with the number of time series. This efficiency arises because the training of \emph{CAND} relies on graphs composed of a fixed number of shapelets, rather than the volume of time series data.

\vspace{1em}
\begin{table}[h]
\small
\centering
\caption{The training time (sec.) of all methods on different pruned datasets.}
\vspace{-1em}
\renewcommand{\arraystretch}{}
\setlength{\tabcolsep}{1.5pt}
\begin{tabular}{l|c|c|c|c}
\toprule[2pt]
Methods & Pruning=0\% & Pruning=30\% & Pruning=50\% & Average\\ \midrule
Naive Bayes~\cite{Watson2001AnES} & 0.004 & 0.004 & 0.003 & 0.004 \\
EE~\cite{Lines2015TimeSC} & 28.833 & 30.771 & 31.358 & 30.321 \\ \midrule
ECTS~\cite{Xing2012EarlyCO} & 19.117 & 14.058 & 10.951 & 14.709 \\ 
MCFEC~\cite{He2015EarlyCO} & 461.811 & 455.344 & 460.894 & 459.350 \\ 
EARLIEST~\cite{Hartvigsen2019AdaptiveHaltingPN} & 18.280 & 15.201 & 13.214 & 15.565\\ 
RHC~\cite{hartvigsen2020recurrent} & 28.175 & 16.536 & 14.202 & 19.638 \\ \midrule
Time2Graph~\cite{cheng2020time2graph} & 97.305 & 84.790 &  77.282 & 86.459\\
AliNet~\cite{sun2020knowledge} & 14.100 & 13.672 & 13.299 & 13.691\\
KE-GCN~\cite{Yu2021KnowledgeEB} & 32.655 & 30.292 & 21.177 & 28.042 \\\midrule
\emph{\textbf{CAND}} & 57.664 & 52.394 &  51.903 & 53.987\\
\bottomrule[1.3pt]
\end{tabular}
\label{table:training_time}
\end{table}

\subsection{Analysis of the Guided Exploration} 

\begin{figure}[h]
\graphicspath{{figs/}}
\begin{center}
\includegraphics[width=0.4\textwidth]{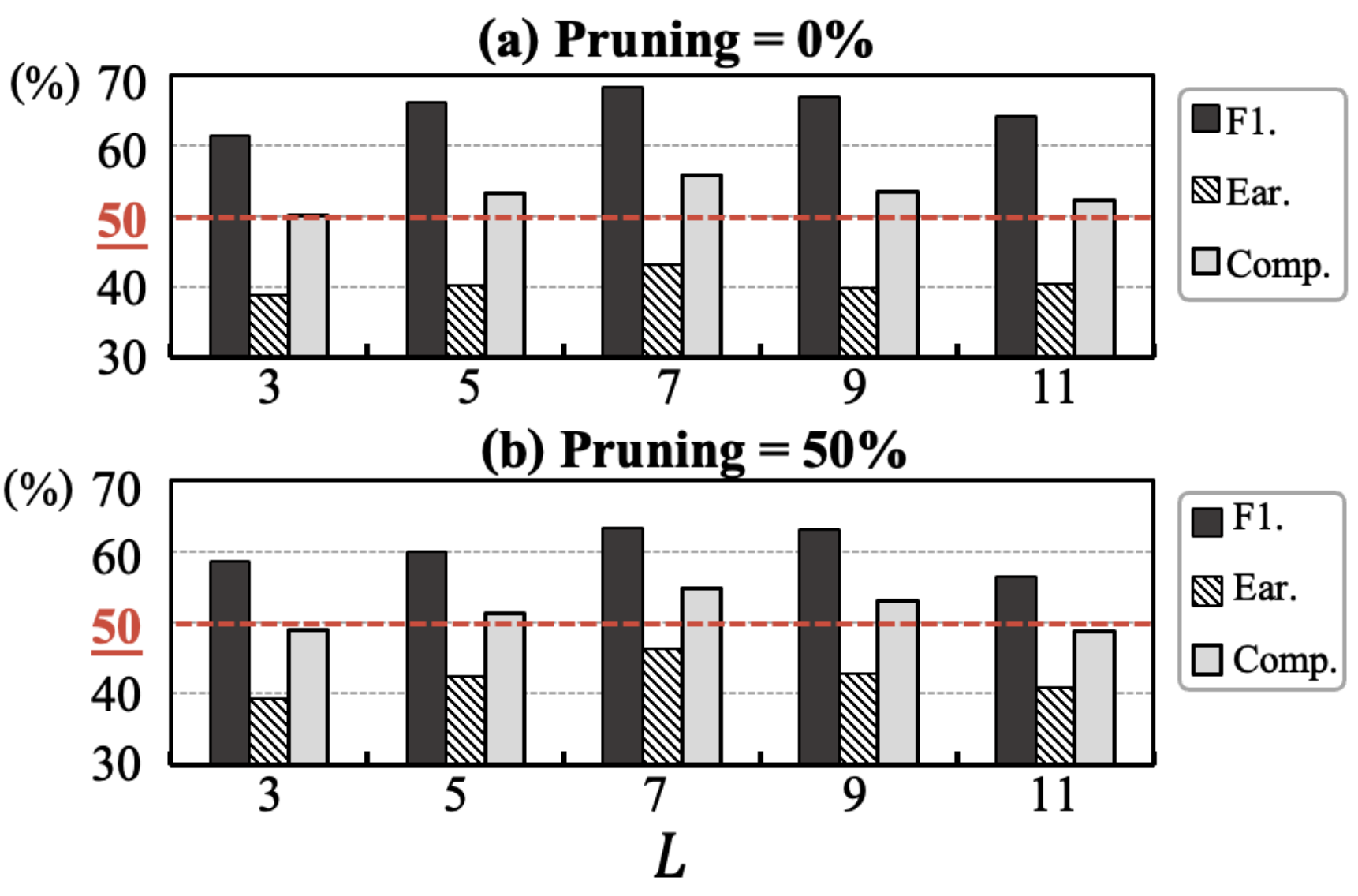}
\end{center}
\vskip -1.2em
\caption{\textcolor{black}{Performance of varying exploration lengths ($L$).}}
\label{fig:exploration_len_prune}
\end{figure}

In our ablation study, we observe that concept-view contextual triplets significantly impact the performance of \emph{CAND}, especially in the 50\% pruned dataset. This leads us to a deeper analysis of these triplets, derived from the designed \textit{guided exploration}, which aims to explore the most influential concepts (most probable transition relationships) for a specific concept. We vary exploration lengths ($L$ in Sec \ref{subsubsec:domain_view}) as $\{3,5,7,9,11\}$ across different pruning levels. The results, shown in Figure~\ref{fig:exploration_len_prune}, indicate that the performance is lowest at $L=3$ and $L=11$, and highest at $L=7$ and $L=9$. This suggests that too short exploration lengths may not capture enough relevant information, while too long lengths may include irrelevant information, as distant transition relationships may not be useful for representing a concept.

\subsection{\textcolor{black}{Analysis of Diverse Patient Populations}}
\label{appendix:population}

In this subsection, we examine the performance of our \emph{CAND} across diverse patient populations.

\subsubsection{Performance across male and female patients}

The performance of \emph{CAND} on different pruned datasets for male and female patients is shown in Table~\ref{table:population_gender}. Our method achieves better accuracy and F1-score for female patients and achieves higher composite scores for female patients in the 30\% and 50\% pruned datasets. While this performance could be attributed to the higher number of male patients in our dataset, it's also possible that differences in vital sign values and sensitivity between males and females~\cite{geovanini2020age, neves2017effect} allow our model to sustain performance for female patients even with less training data.

\vspace{-1em}
\subsubsection{Performance across different age groups}
The performance of \emph{CAND} across four age groups—0-40, 40-60, 60-80, and above 80 (denoted as ``80 $+$'')—is presented in Table~\ref{table:population_age}. It can be observed that the composite score for patients older than 80 years is, on average, lower compared to the other age groups. This could be due to typically less pronounced changes in vital signs among the elderly~\cite{chester2011vital}, which may result in lower detection capabilities for deterioration by the model. However, on average, the differences in composite scores among the 0-40, 40-60, and 60-80 age groups are relatively minor. Although the earliness score for the 0-40 age group is lower, its accuracy is the highest among all age groups, indicating that our method still performs effectively for patients under 80 years of age.

\begin{table}[t]
\small
\centering
\caption{Comparison of performance (in \%) between male and female patients.}
\vspace{-1em}
\renewcommand{\arraystretch}{1}
\setlength{\tabcolsep}{2pt}
\begin{tabular}{lc|ccc|c}
\toprule[1.3pt]
\multicolumn{2}{c|}{Patient Populations} &  Accuracy & F1-score & Earliness & Composite Score \\ \midrule
\multirow{2}{*}{Pruning=0\%} & Male & 55.67 & 67.29 & 45.64 & 56.46 \\
 & Female & 63.15 & 70.78 & 38.29 & 54.53 \\ \midrule
\multirow{2}{*}{Pruning=30\%} & Male & 54.45 & 64.57 & 47.66 & 56.11 \\
 & Female & 55.26 & 67.81 & 46.94 & 57.37 \\\midrule
\multirow{2}{*}{Pruning=50\%} & Male & 54.45 & 59.81 & 46.55 & 53.18 \\
 & Female & 71.05 & 71.42 & 46.52 & 58.97 \\ \midrule\midrule
 \multirow{2}{*}{Average} & Male & 54.85 & 63.89 & 46.61 & 55.25 \\
 & Female & 63.15 & 70.00 & 43.91 & 56.95 \\
\bottomrule[1.3pt]
\end{tabular}
\label{table:population_gender}
\end{table}

\begin{table}[t]
\small
\centering
\caption{Comparison of performance (in \%) between different ages of patients. }
\vspace{-1em}
\renewcommand{\arraystretch}{1}
\setlength{\tabcolsep}{2pt}
\begin{tabular}{lc|ccc|c}
\toprule[1.3pt]
\multicolumn{2}{c|}{Patient Populations} &  Accuracy & F1-score & Earliness & Composite Score \\ \midrule
\multirow{4}{*}{Pruning=0\%} & 0-40 & 70.00 & 78.57 & 40.00 & 59.28 \\
 & 40-60 & 59.28 & 68.90 & 46.27 & 57.59 \\ 
& 60-80 & 55.20 & 64.36 & 44.25 & 54.30 \\ 
& 80 $+$ & 53.33 & 60.00 & 28.12 & 44.06 \\ 
 \midrule
\multirow{4}{*}{Pruning=30\%} & 0-40 & 70.00 & 78.57 & 39.68 & 59.12 \\
 & 40-60 & 53.40 & 64.36 & 52.23 & 58.29 \\ 
& 60-80 & 50.00 & 61.25 & 47.10 & 54.17 \\ 
& 80 $+$ & 63.33 & 65.00 & 51.56 & 58.28 \\ 
 \midrule
\multirow{4}{*}{Pruning=50\%} & 0-40 & 63.33 & 65.00 & 37.50 & 51.25 \\
 & 40-60 & 59.28 & 63.80 & 53.41 & 58.60 \\ 
& 60-80 & 58.33 & 59.80 & 42.61 & 51.20 \\ 
& 80 $+$ & 64.44 & 46.15 & 30.72 & 38.43 \\ \midrule\midrule
\multirow{4}{*}{Average} & 0-40 & 67.77 & 74.04 & 39.06 & 56.55 \\
 & 40-60 & 57.32 & 65.68 & 50.63 & 58.16 \\ 
& 60-80 & 54.51 & 61.80 & 44.65 & 53.22 \\ 
& 80 $+$ & 60.36 & 57.05 & 36.80 & 46.92 \\ 
\bottomrule[1.3pt]
\end{tabular}
\label{table:population_age}
\end{table}

\subsection{\textcolor{black}{Performance across Different Levels of Patient Deterioration}}
\label{appendix:different_deter}

To recognize different levels of illness deterioration, we divided the testing data based on whether patients' Apache II scores increased or decreased by 0 to 5, 5 to 10, and more than 10 points (denoted as ``10 $\uparrow$''). This allowed us to assess the performance of our model, \emph{CAND}, across these ranges, as illustrated in Table~\ref{table:range_score}. Across all pruned datasets, our model demonstrates higher accuracy for patients experiencing significant changes in Apache II scores (increases or decreases of up to 10 points). This is intuitive, as larger shifts in patient conditions often lead to more noticeable changes in vital signs, making it easier for the model to identify these changes and reduce false positive rates. Notably, for patients experiencing minor changes in Apache II scores (0-5 points), the highest composite score is achieved on the average performance of all pruned datasets. These results underscore the capability of \emph{CAND} to effectively detect significant changes in patient conditions while adeptly balancing earliness and detection performance among patients with subtle changes.

\begin{table}[t]
\small
\centering
\caption{Comparison of performance (in \%) between different ranges of changes in the Apache II scores. }
\vspace{-1em}
\renewcommand{\arraystretch}{1}
\setlength{\tabcolsep}{2pt}
\begin{tabular}{lc|ccc|c}
\toprule[1.3pt]
\multicolumn{2}{c|}{Patient Populations} &  Accuracy & F1-score & Earliness & Composite Score \\ \midrule
\multirow{3}{*}{Pruning=0\%} & 0-5 & 48.41 & 60.06 & 36.87 & 48.46 \\
 & 5-10 & 52.25 & 60.39 & 41.21 & 50.80 \\ 
& 10 $\uparrow$ & 84.61 & 88.88 & 38.68 & 63.78  \\ 
 \midrule
\multirow{3}{*}{Pruning=30\%} & 0-5 & 39.87 & 54.00 & 45.78 & 49.89\\
 & 5-10 & 51.90 & 58.33 & 46.83 & 52.58 \\ 
& 10 $\uparrow$ & 75.96 & 84.03 & 36.75 & 60.39 \\ 
 \midrule
\multirow{3}{*}{Pruning=50\%} & 0-5 & 61.90 & 67.61 & 63.28 & 65.44\\
 & 5-10 & 51.56 & 53.58 & 41.87 & 47.73\\ 
& 10 $\uparrow$ & 72.11 & 73.33 & 45.83 & 59.58 \\  \midrule\midrule
\multirow{3}{*}{Average} & 0-5 & 50.06 & 60.55 & 48.64 & 54.60\\
 & 5-10 & 51.90 & 57.43 & 43.30 & 50.37 \\ 
& 10 $\uparrow$ & 77.56 & 82.08 & 40.42 & 61.25 \\ 
\bottomrule[1.3pt]
\end{tabular}
\label{table:range_score}
\end{table}

\subsection{\textcolor{black}{Supplementary Discussion for the AUC performance of \emph{CAND}.}}
\label{appendix:observ_length}

In previous experiments (Table~\ref{table:comparison}), although the AUC value of \emph{CAND} achieve the best among all methods, the AUC value is not particularly high (around 62\%) due to two reasons:

\begin{itemize}[leftmargin=*]
    \item Instead of analyzing patients' complete electronic health records, such as demographics, comorbidities, and lab test results, our work focuses on monitoring only vital signs, which are simple and cost-effective to collect. Also, monitoring vital signs is a non-invasive method that can reduce the risk of nosocomial infections~\cite{sikora2020nosocomial}. Therefore, our work can be considered an information-scarce scenario as mentioned in Sec.~\ref{subsec:comparison}.
    \item In addition, \emph{CAND} aims to obtain correct detection results as early as possible to balance detection earliness and reliability. This approach may cause the model to stop monitoring early and decide on the detection result, which may limit detection reliability (accuracy, F1-score, recall, and AUC value).
\end{itemize}

To examine the extent to which detection reliability can be achieved while maintaining an acceptable earliness score, we conduct a simple experiment to enforce \emph{CAND} to start detecting deterioration only after observing the initial 50\% the total vital sign time series, as shown in Table~\ref{table:start_from_length}. Additionally, we analyze the performance after observing at least 50\% of the total vital sign time series among patients with different Apache II score changes (0-5 and 5 $\uparrow$) to represent two groups of patients at different levels of deterioration and recovery, as revealed in Table~\ref{table:length_50_level}.

From Table~\ref{table:start_from_length},  it is evident that the later the detection starts, the higher the accuracy and AUC. Although recall decreases slightly, the increase in precision ensures that the F1-score also increases.
Table~\ref{table:length_50_level} reveals that after observing 50\% of vital signs, patients with changes in Apache II scores exceeding 5 points (5 $\uparrow$) experienced a lower earliness score due to the later start in detection, yet they achieved good results in accuracy, recall, F1, and AUC. Patients with Apache II score changes between 0-5 points, while showing lower accuracy, recall, and F1-score compared to the overall patients, still reached an AUC of over 0.70. These preliminary experimental adjustments suggest that reducing the emphasis on earliness can enhance detection reliability. This also demonstrates that our method, \emph{CAND}, is applicable in scenarios with varying importance on detection earliness.

\begin{table}[h]
\small
\centering
\caption{Performance (in \%) of detection starting after observing different lengths of vital signs in the 0 \% pruned dataset.}
\vspace{-1em}
\renewcommand{\arraystretch}{}
\setlength{\tabcolsep}{1.5pt}
\begin{tabular}{l|ccccc}
\toprule[2pt]
 & Acc. & Rec. & F1. & AUC & Ear. \\ \midrule
Detection after 0\% of vital signs & 58.11 & 92.86 & 68.39 & 62.25 & 43.08 \\
Detection after 50\% of vital signs & 67.98 & 90.59 & 73.33 & 72.76 & 20.67 \\
\bottomrule[1.3pt]
\end{tabular}
\label{table:start_from_length}
\end{table}

\begin{table}[h]
\small
\centering
\caption{Performance (in \%) across different ranges of changes in Apache II scores, with detection starting after observing 50\% of total vital sign time series.}
\vspace{-1em}
\renewcommand{\arraystretch}{}
\setlength{\tabcolsep}{1.5pt}
\begin{tabular}{l|ccccc}
\toprule[2pt]
Ranges of changes & Acc. & Rec. & F1. & AUC & Ear. \\ \midrule
Overall & 67.98	& 90.59	& 73.33 & 72.76 & 20.67 \\
0-5 & 61.66	& 81.94	& 69.99	& 70.71	& 19.79 \\
5 $\uparrow$ & 70.70 & 96.42 &	74.21 & 75.54 & 16.51 \\
\bottomrule[1.3pt]
\end{tabular}
\label{table:length_50_level}
\end{table}

\subsection{Supplementary Results for Parameter Sensitivity Analysis} 
\label{appendix:param}

\begin{figure}[h]
\graphicspath{{figs/}}
\subfloat[F1-score.] {
\label{fig:param_cont_weight_f1}
\includegraphics[width=0.22\textwidth]{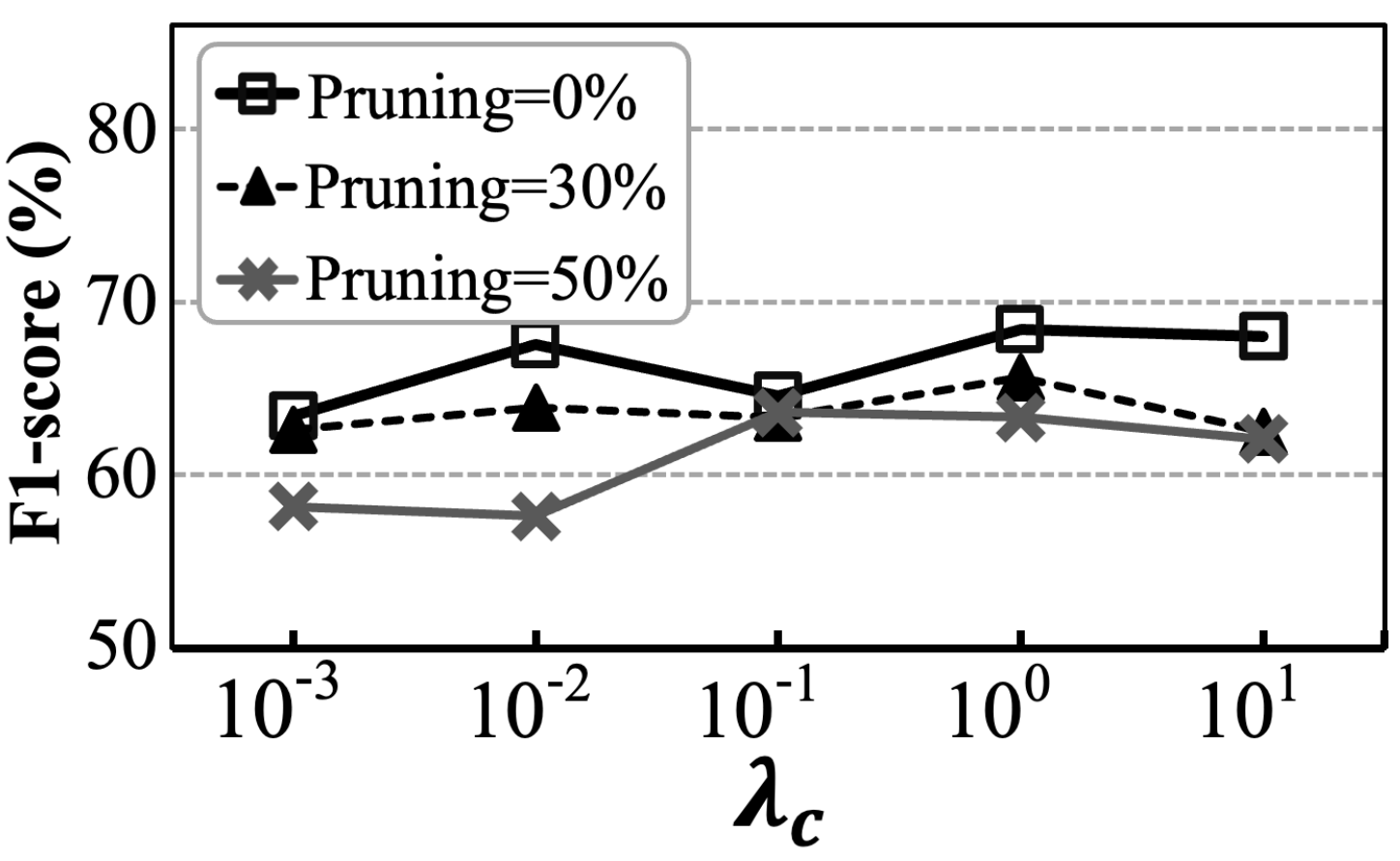}
}
\subfloat[Earliness Score.] {
\label{fig:param_cont_weight_ear}
\includegraphics[width=0.22\textwidth]{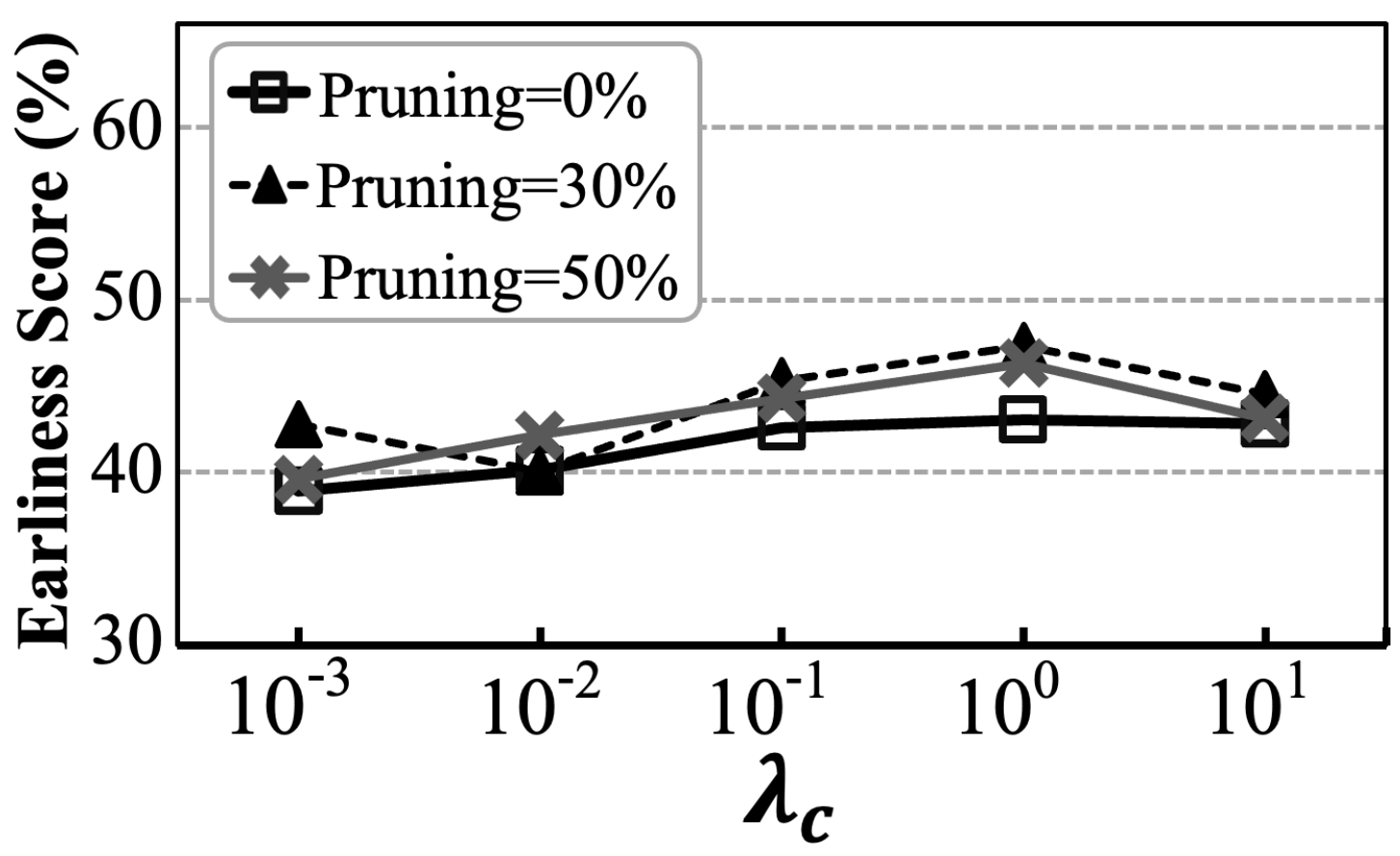}
}
\vspace{-1.2em}
\caption{\textcolor{black}{F1-scores and earliness scores of \emph{CAND} with different weights of contextual loss ($\lambda_c$)}}%
\label{fig:param_cont_weight_f1_ear}%
\end{figure}

\begin{figure}[h]
\graphicspath{{figs/}}
\vspace{-1em}
\subfloat[F1-score.] {
\label{fig:param_scale_f1}
\includegraphics[width=0.23\textwidth]{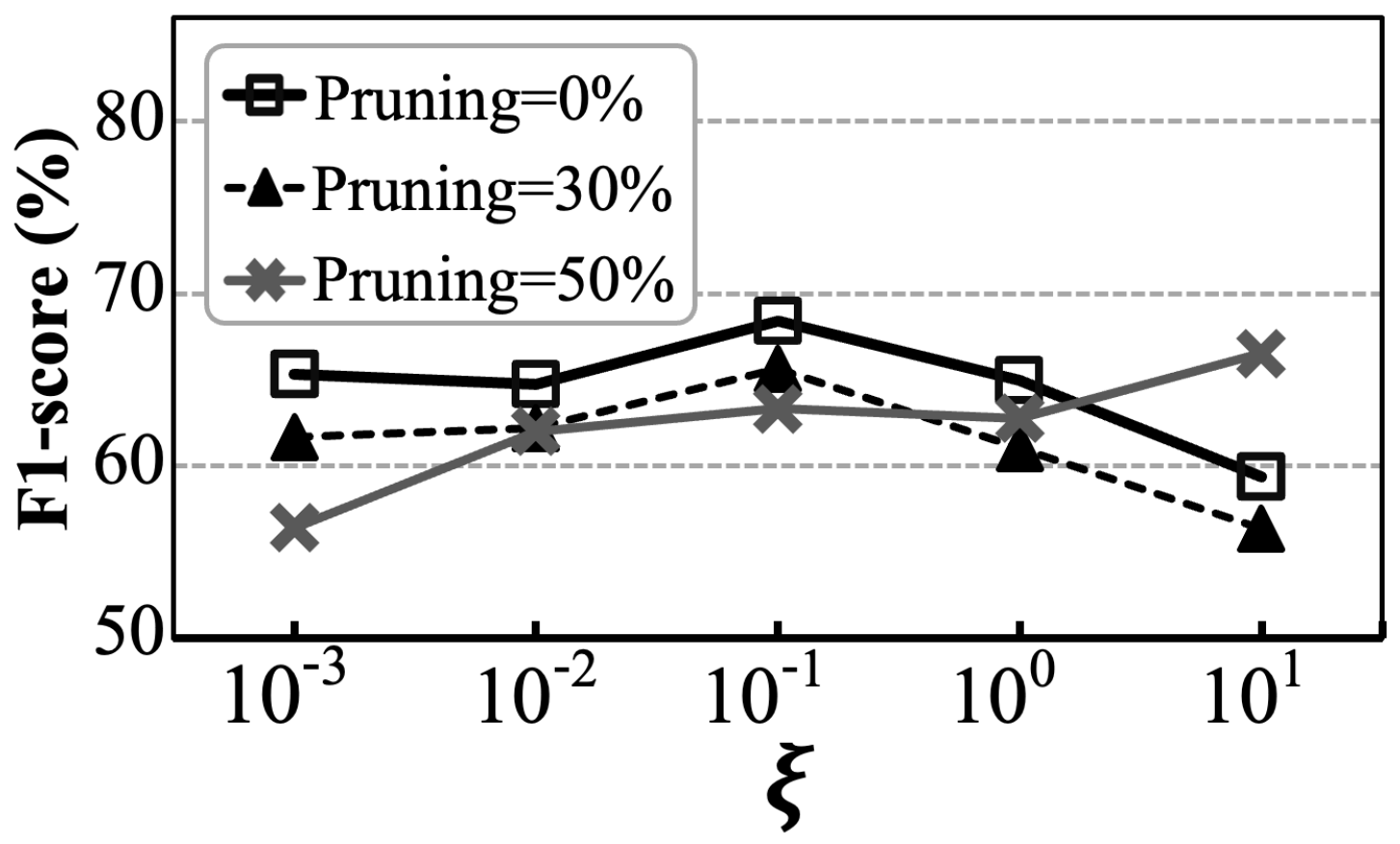}
}
\subfloat[Earliness Score.] {
\label{fig:param_scale_ear}
\includegraphics[width=0.23\textwidth]{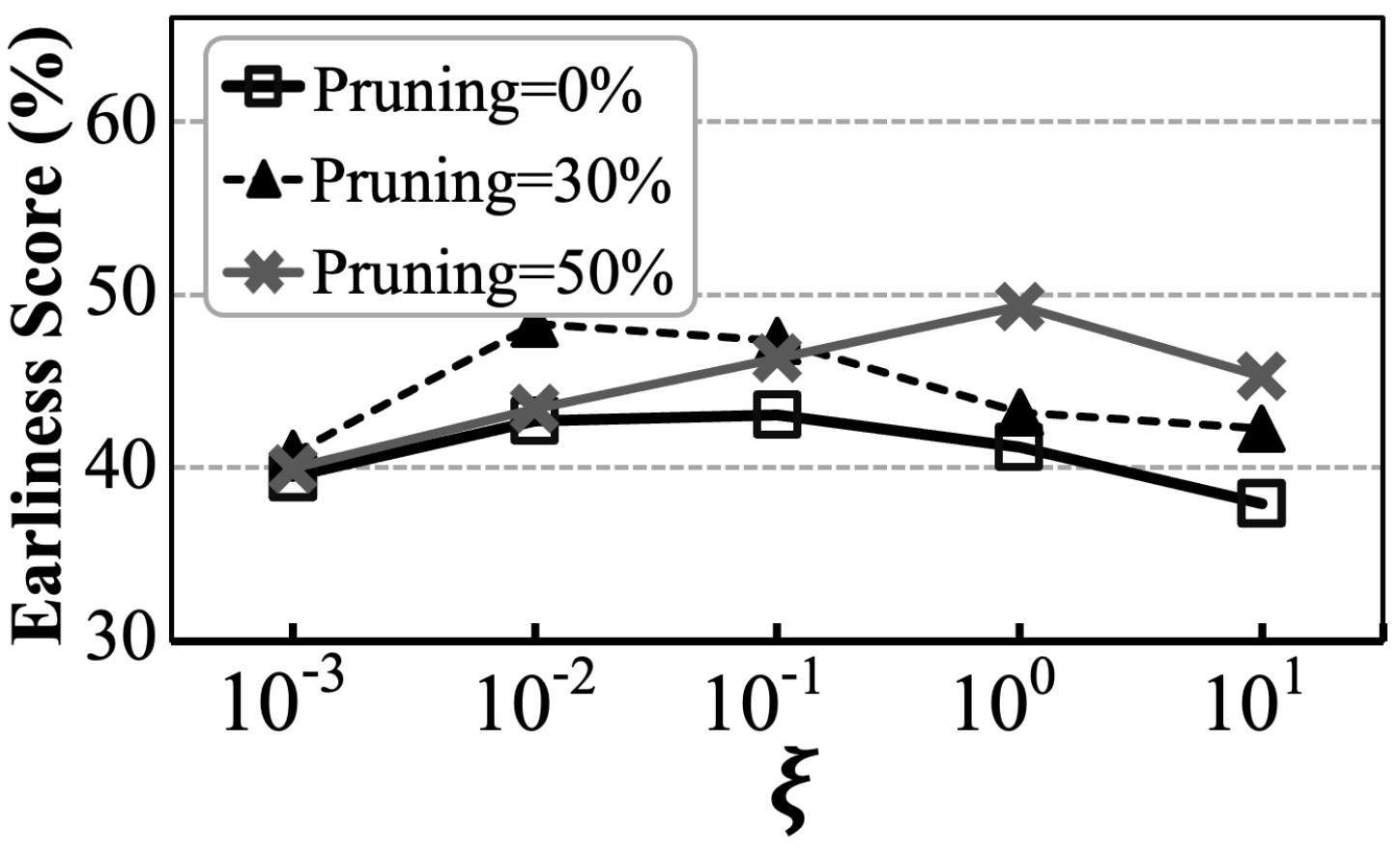}
}
\vspace{-1em}
\caption{\textcolor{black}{F1-scores and earliness scores of \emph{CAND} with different values of the scale factor ($\xi$).}}%
\label{fig:param_scale_f1_ear}%
\end{figure}

In this section, we supplement our earlier parameter sensitivity analysis (Sec.\ref{subsec:param}) with F1-score and earliness score results. Figures~\ref{fig:param_cont_weight_f1_ear} and \ref{fig:param_scale_f1_ear} display outcomes for variations in contextual loss weight ($\lambda_c$) and scale factor ($\xi$), respectively. Consistent with findings in Sec.\ref{subsec:param}, F1-score and earliness score of \emph{CAND} improve as $\lambda_c$ increases from $10^{-2}$ to $10^0$ in the 30\% and 50\% pruned datasets, but decrease at $10^1$. This suggests that excessively high weights on contextual triplets diminish the significance of original triplets in cross KS, degrading representation quality. For $\xi$, optimal results in 0\% and 30\% pruned datasets occur at $10^{-1}$ or $10^{-2}$, highlighting the inefficacy of extreme margin adjustments.

\subsection{\textcolor{black}{Further Discussion}} 
\label{appendix:discuss}

While our \emph{CAND} model effectively balances detection earliness and effectiveness as shown in Table~\ref{table:comparison}, enhancing its accuracy remains a challenge due to its sole reliance on vital sign data. Vital signs are significantly influenced by individual patient characteristics, including gender, age, and BMI. Future enhancements could focus on integrating these factors and exploring their causal relationships with vital sign dynamics. This could lead to a more nuanced understanding of illness deterioration signs and potentially improve the accuracy of CAND across diverse patient populations.

Additionally, while in ICUs, the earliness of detection is as crucial as the reliability of the results (e.g., accuracy, recall, F1-score), the importance placed on detection earliness versus reliability may vary in different scenarios. For example, in non-acute wards, where patient conditions are generally less urgent, earliness might be less critical. Investigating how the model could automatically adjust its decision-making process for patients with varying degrees of urgency could be a promising direction for future research. Moreover, the dataset currently in use involves a limited number of patients. One of our future objectives is to actively acquire more diverse patient data to enhance the generalizability of our model.

\end{document}